\documentclass[conference]{IEEEtran}
\usepackage{times}

\usepackage[numbers]{natbib}
\usepackage{multicol}
\usepackage{graphics,graphicx,caption,float,subcaption,booktabs,xcolor,multirow,array,color,ifthen,tabu,colortbl,dblfloatfix,url,xparse,mathtools,patchcmd,algorithm,algorithmic,amssymb,xspace,nicefrac,microtype,amsmath,amsfonts,bm,ragged2e,tikz,stackengine,etoolbox,xpatch,enumerate,xstring,setspace,tabularx,makecell,changepage,cuted,titlesec,wrapfig,tcolorbox, sidecap}
\usepackage[pagebackref=true,breaklinks=true,colorlinks=true,bookmarks=false,citecolor=blue]{hyperref}

\begin{document}
\newcommand{\DP}[1]{{\color{red} {DP: {#1}}}}
\newcommand{\aravind}[1]{{\color{blue} {Aravind: {#1}}}}
\newcommand{\our}{H-NDP\xspace}
\newcommand{\ours}{H-NDPs\xspace}

\title{Hierarchical Neural Dynamic Policies}
\author{Shikhar Bahl$\qquad$Abhinav Gupta$\qquad$Deepak Pathak\\Carnegie Mellon University}

\maketitle

\begin{abstract}
We tackle the problem of generalization to unseen configurations for dynamic tasks in the real world while learning from high-dimensional image input. The family of nonlinear dynamical system-based methods have successfully demonstrated dynamic robot behaviors but have difficulty in generalizing to unseen configurations as well as learning from image inputs. Recent works approach this issue by using deep network policies and reparameterize actions to embed the structure of dynamical systems but still struggle in domains with diverse configurations of image goals, and hence, find it difficult to generalize. In this paper, we address this dichotomy by leveraging embedding the structure of dynamical systems in a hierarchical deep policy learning framework, called Hierarchical Neural Dynamical Policies (H-NDPs). Instead of fitting deep dynamical systems to diverse data directly, H-NDPs form a curriculum by learning local dynamical system-based policies on small regions in state-space and then distill them into a global dynamical system-based policy that operates only from high-dimensional images. H-NDPs additionally provide smooth trajectories, a strong safety benefit in the real world. We perform extensive experiments on dynamic tasks both in the real world (digit writing, scooping, and pouring) and simulation (catching, throwing, picking). We show that H-NDPs are easily integrated with both imitation as well as reinforcement learning setups and achieve state-of-the-art results. Video results are at~\url{https://shikharbahl.github.io/hierarchical-ndps/}.

\end{abstract}

\IEEEpeerreviewmaketitle

\section{Introduction}
\begin{figure}[t]
  \centering
  \includegraphics[width=\linewidth]{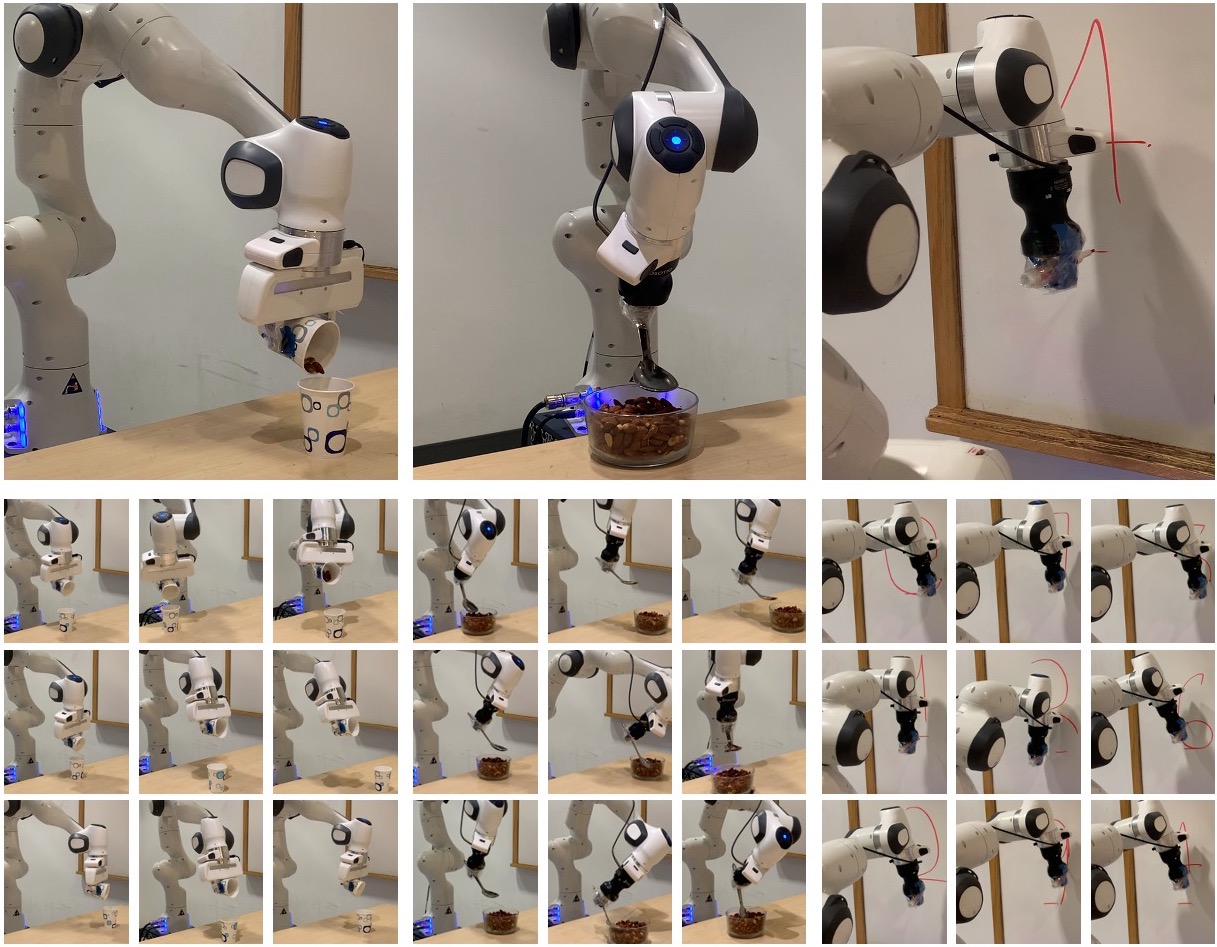}
  \caption{\small We present \ours, an efficient real-world robot learning algorithm. Our method is able to perform scooping, writing and pouring from image input only. We are able to generalize across a high amount of diversity, i.e. different object positions, pose, etc. Videos at~\url{https://shikharbahl.github.io/hierarchical-ndps/}.}
  \label{fig:teaser}
\end{figure}

Consider the tasks such as pouring liquid or scooping beans as shown in Figure~\ref{fig:teaser}. These tasks are dynamic in nature, i.e., they require the robot to continuously apply the right forces and accelerations to act in a reactive manner to changes in the environment. Unlike quasi-static tasks, e.g. pushing or 2D grasping, where it can take arbitrarily long in between each action, the robot needs to reason at the whole trajectory level to execute a swift motion to perform dynamic tasks. For instance, if the robot scoops the beans too slowly they will fall back into the bowl, or if scooped too quickly the beans will be thrown out of the bowl. A common way to address this trajectory-level reasoning is to encode robot movements using nonlinear dynamical systems, like the ones that govern the flow of heat or the movement of planets. This idea is encapsulated by a family of methods known as Dynamic Movement Primitives (DMPs)~\cite{schaal2006dynamic, isprt2012dmp}, which can compactly represent basic building block trajectories that are then stitched together to perform complex tasks. DMPs restrict the space of permissible robot movements by constraining the robot's goal and trajectory shape to obey a parametric nonlinear differential equation, consistent with the robot's kinematics and dynamics. This allows DMPs to effectively reason in the space of entire \textit{trajectories}, rather than at the level of individual actions. Consequently, these approaches have led to impressive demos such as pancake flipping~\cite{kormushev2010robot}, dart throwing~\cite{kober2011reinforcement} or playing table tennis~\cite{muelling2013tabletennis}.  However, this class of approaches suffers from two drawbacks -- (a) Generalization: Because of the constraints they impose on trajectories, DMPs do not have the power to represent general movements, and limit the controller to small variations in the initial or goal states. (b) Image-Observations: DMPs have mostly been built on estimating state vectors and struggle with high-dimension input, such as raw images. These shortcomings are in contrast to the flexibility allowed by deep learning methods in terms of generalization to unseen scenarios and scalability to high-dimensional image inputs~\cite{levineFDA15,pinto2015supersizing,mahler2017dex,kalashnikov2018qt}. However, most of the real robot results using deep learning methods are still limited to pick and place-style quasi-static tasks, as compared to the dynamic tasks achieved by DMP-based methods. 

In this paper, our goal is to address the following question: can we build \textit{generalizable} robot policies for dynamic tasks by combining the ability of DMPs to reason in the space of trajectory distributions, with the ability of modern deep robot learning methods to learn from unstructured image data?  Recent works~\cite{bahl2020neural, pahic2018deepenc} address this via Neural Dynamic Policies (NDPs)~\cite{bahl2020neural} by embedding structure of a nonlinear dynamical system as a differentiable layer within the policy network and training end-to-end. However, similar to DMPs, these methods still struggle to generalize to unseen state configurations. Why is that? Consider the examples in Figure~\ref{fig:teaser} where we see a large diversity in the location and pose of objects. Depending on the initial state of the robot and the location of the containers, the joint trajectories will need to be diverse enough in terms of reachability and dynamics, to successfully perform the task of pouring or scooping. Hence, a direct attempt to fit a single policy to all these trajectories poses a big practical challenge in optimization which is further aggravated when the input is a high-dimensional image -- which we argue is the case in most interesting real-world robotic problems. 

Our solution is to harness the individual strengths of both dynamical systems for movement representations and deep network policies.
Instead of directly fitting a single policy on diverse scenarios, we fit dynamical system-based policies first in local regions of the task space and then distill them together into a global one.

Our system consists of a library of \textit{local} NDPs~\cite{bahl2020neural} and a single \textit{global} NDP. Each local NDP exploits the strengths of DMPs: (a) overfit to operate in small regions of the task space and ensure task success at all times; (b) operate on privileged low-level state information as input. The global NDP is meant to operate on the entire space and only receives raw sensory data as input, e.g., raw images. This global policy is trained to not only maximize the task performance but also to imitate the behavior of the local policies. The key here is that both local and global policies have different objectives: local policies place importance on task success in their local regions, and the global policy places importance on learning from images in a generalizable manner. Owing to this local-to-global structure, we call our framework \textit{Hierarchical Neural Dynamic Policies (\ours)}.

\begin{figure*}[t]
  \centering
  \includegraphics[width=0.9\linewidth]{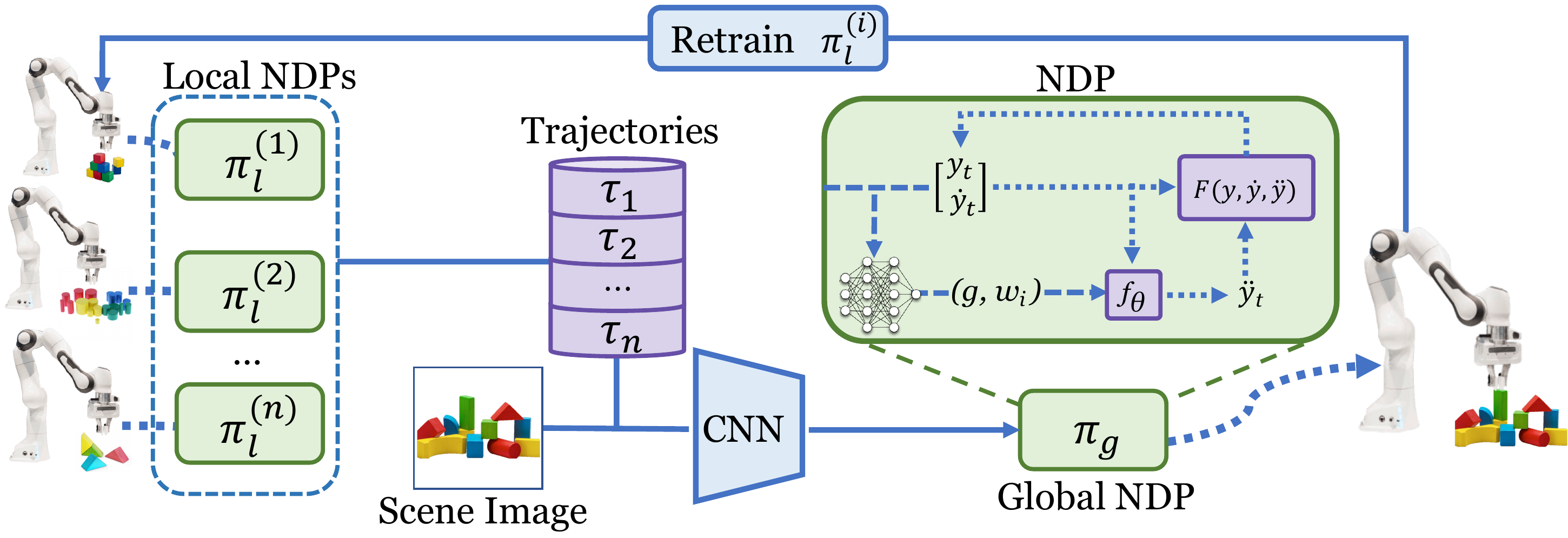}
  \caption{\small We train local Neural Dynamic Policies (NDPs) $\pi_{l}^{(i)}$ on each region $i$ of the task space, from state observations. A global NDP $\pi_g$ (usually taking in image input $I_t$) learns to imitate the local experts. We use the global NDP to retrain local NDPs which keeps the NDPs from diverging. These local-to-gloabl interactions happen in an iterative manner. NDPs make a good candidate for capturing such local-to-global interactions due to their shared structure and the fact that they operate over a smooth trajectory space.}
  \vspace{-0.1in}
  \label{fig:method}
\end{figure*}

Training \ours for real-world robot learning comes with a practical challenge: there is no guarantee that the local trajectories distilled by the global NDP will indeed be successful when tried even on the same local locations as training unless it is completely overfitting in which case it will not generalize to new locations. Instead of hoping it to just work, we perform multiple iterations of this local-to-global procedure by re-training local NDPs by solving local tasks while being faithful to the global NDPs and then distilling the refined local ones into a new global NDP until convergence, as shown in Figure~\ref{fig:method}. Such an iterative process is standard practice in general~\cite{levine2013guided,bucilua2006model,rusu2015policy} to prevent the distilled network from diverging. However, the added advantage of \ours is that the embedded dynamical system enhances both safety, convergence, and overall performance, as we show in the results section later.

One of the big contributions of this paper is an exhaustive experimental evaluation of \ours and several other baselines on real-world tasks of writing, scooping, and pouring with a robot. Our real-world experiments are conducted in realistic settings with raw high-dimensional images as inputs, with large variations in object positions and goal locations, involving several hundred hours of robot interaction. Finally, we also evaluate complex simulated tasks like throwing, catching, and picking. We show that \ours achieve state-of-the-art performance across all the tasks in reinforcement as well as imitation learning settings.


\section{Background}
\subsection{Dynamical Systems in Robotics}
\label{sec:dmp}
Dynamical systems have long been used in robotics to represent trajectories and various motion primitives. Such systems operate over an arbitrary robot state (say $y$, $\dot{y}$ and $\ddot{y}$). Examples of such coordinate systems are joint angles or end-effector positions. Specifically, past work has used a second order dynamical system called Dynamic Movement Primitives (DMPs) ~\cite{isprt2012dmp,schaal2006dynamic,pastor2009motorskills}, derived from Lagrangian Mechanics, to represent robot motions. DMPs are represented by the following: 

\begin{equation}
    \ddot{y} = \alpha(\beta(g - y) - \dot{y}) + f(x, g),
    \label{eq:dmp}
\end{equation}

Here, $g$ is a desired goal state and $\alpha$ is a hyperparameter (and $\beta = \frac{\alpha}{4}$, in order for critical damping). The above equation can be broken down into two parts. $\alpha(\beta(g - y) - \dot{y})$ allows for smooth convergence to the goal, making the trajectory physically realizable. $f(x, g)$, a nonlinear forcing function, captures the shape of the trajectory. It is a common practice to use radial basis functions to represent $f$; the combination of these allows $f$ to represent arbitrary shapes. Traditionally, the weights on these basis functions, $w_i$, are fit via regression on demonstration trajectories.    
 \begin{equation}
    f(x, g) = \frac{\sum \psi_i w_i}{\sum \psi_i}x(g - y_0), \quad
    \psi_i = e^{(-h_i(x-c_i)^2)}
    \label{eq:forcing_func}
\end{equation}
$f$ decays linearly with $x$, which is a variable used to replace the time dependency of this ODE. It allows us to arbitrarily stretch or compress time, and sample trajectories of any length from the DMP. $x$ obeys the following first order dynamical system: $\dot{x} = -\alpha_x x$

\subsection{Neural Dynamic Policies}
\label{sec:ndps}
More recently, DMPs have been used in a deep policy learning setup \cite{bahl2020neural, pahic2018deepenc}. Neural Dynamic Policies (NDPs) \cite{bahl2020neural} embed the dynamical system described in DMPs inside a deep policy network. Given an input (image or state), $s_t$, NDPs employ a neural network $\Phi$ and output DMP parameters, $w_1, ..., w_n$ (radial basis function weights) and goal state $g$. These parameters are used by a forward integrator to output a trajectory $\{y_t,\dot{y_t},\ddot{y_t}\}$ for $t=1$ to $t=T$. If the robot's action space is not in the same coordinate system as $y$, then an inverse controller $\Omega(.)$ is used to convert the desired trajectory into a set of actions for the robot to execute in the environment. The forward pass of an NDP involves solving a second order ODE and a pass through the inverse controller. \citet{bahl2020neural} show that NDPs are fully differentiable and can be efficiently incorporated in RL or Imitation Learning settings, and demonstrate some toy results showing that NDPs can learn from images as well.


\section{Method}
\label{sec:method}
Consider the real world task of scooping from a bowl. The robot has to both plan a trajectory that will allow it to do the scooping motion properly, and understand any potential randomness, for instance if the bowl changes locations. A single policy likely will have a lot of trouble with this. We address such challenges by presenting Hierarchical Neural Dynamic Policies (\ours). \ours use a local-to-global learning scheme which makes it much easier for the agent to learn how to handle diversity in the task and deal with raw image inputs. We leverage structure provided by NDPs \cite{bahl2020neural} for policy learning, allowing our hierarchical policies to operate in a shared space, and thus leads to smoother trajectories and more sample efficient learning. In this section, we describe how this setup works, both in the reinforcement and imitation learning settings.

\subsection{Hierarchical Neural Dynamical Policies}
 We break down policy learning for a given task into two components: local controllers which operate from exact state observations, and a global policy which learns from raw sensory observations, for example robot poses and images. Both policies are NDP; the policy networks have an embedded dynamical system as a layer. Directly optimizing the global policy for the full task can be difficult, since dynamical systems can easily overfit to a single trajectory. Let $\pi$ be an NDP and let $\phi(s; \theta)$ be the deep network inside the NDP, parameterized by weights $\theta$. $\phi(s; \theta)$ outputs the DMP parameters which are used by the forward integrator $F$ to solve the differential equation described in Equation~\ref{eq:dmp}.  $\tau$, the output of $\pi(.|s)$, is $F(\phi(s; \theta))$. 
 
We divide the task space into $M$ different regions. We train local NDPs $\pi_{l}^{(i)}$ on each region $i$ ($R_i$) of the task space. For example, for a task like scooping, each bowl location would be its own region, and we would train a single NDP to solve the task for that specific bowl location. For the rest of this section, let the local policy $\pi_{l}^{(i)}$ be parameterized by network weights by $\theta_{l}^{(i)}$. For task $i$, we will compute loss on the NDP output,  $\mathcal{L}_i(\theta_l^{(i)})$ and optimize with respect to $\theta_{l}^{(i)}$. This loss can be any differentiable loss based on the policy output. In the next two sections, we will describe what $\mathcal{L}_i$ is in the case of RL and imitiation learning.  

Once the local policies are trained, the global NDP $\pi_g$ parameterized by network weights $\theta_{g}$, learns to imitate the local policies. This makes it easier for the global policy to understand the difference between high level task details and low level task optimization. The global NDP conditioned on the current observation, $s_t$, learns to clone the behavior of the local NDPs in using the loss: $\mathcal{L}_\text{BC} = \sum_{i} ||F(\phi(s_t; \theta_g)) -  F(\phi(s_t; \theta_l^{(i)}))||_2$. There is no guarantee, however, that a single iteration of behavior cloning will work. In practice, an iterative process is standard. Therefore, we fine-tune $\pi_g$ on the loss function for the full task (union of all task regions $R_i$), $\mathcal{L}_g(\theta_g)$. In summary, the overall global NDP training loss is:
\begin{equation}
    \label{eq:global-loss}
     \mathcal{L}_{\text{global}} = \mathcal{L}_\text{BC} + \mathcal{L}_g(\theta_g)
\end{equation}
We would like to minimize the amount of human supervision, thus we do not want to create more task spaces but need more data to train the policy. One possibility is to retrain the local NDPs and collect more data, However this could easily lead to divergence. Thus we use the NDP $\pi_g$ to reduce divergence by adding $\alpha_i D_{KL}(\pi_{l}^{(i)} || \pi_g)$ to the local NDP task loss $\mathcal{L}_i(\theta_l^{(i)})$. $\alpha_i$ is a hyperparameter for the weight of this extra loss term. We collect more data from the local experts and repeat the above steps until convergence. We call this process \textit{iterative refinement}. This structure allows the local experts to adjust their outputs based on what is easier for the global policy to learn. NDPs make a good candidate for such a learning scheme due to their shared structure and the fact that they both operate over a smooth trajectory space, leading to much more efficient learning. Hence, the overall loss for the $i$th local NDPs is: 
\begin{equation}
    \label{eq:local-loss}
    \mathcal{L}_{\text{local}} = \mathcal{L}_i(\theta_l^{(i)}) + \alpha_i D_{KL}(\pi_{l}^{(i)} || \pi_g)
\end{equation}
We provide a detailed description of \ours in Algorithm 1.  

The general idea of local-to-global learning has been widely studied in machine learning for generalization in complex domains~\cite{sun2001implicit,bucilua2006model,rusu2015policy}. For robot learning in particular, local-to-global structure has been exploited by for imitation learning by~\citet{levine2013guided} and for RL by~\citet{ghosh2017divide, teh2017distral}. In contrast to the black-box policy networks used in these works, our main contribution is to embed the structure of a nonlinear dynamical system within the network. NDPs make the interactions between global and local policies a lot more efficient, since both operate in the same DMP parameter space. This allows generalization to new configurations for dynamic tasks, a strong advantage of our method. We now discuss how to apply \ours to both imitation and RL settings in the following subsections.

\subsection{\ours for Imitation Learning}
In the imitation learning setting, we train the global NDP $(\pi_g)$ via visual inputs and the local NDPs ($\pi_l^{(i)}$) are trained via supervised learning to imitate kinesthetic demonstrations. We start with a single demonstration for each $R_i$. Let this demonstration be $\tau_\text{demo}^{(i)}$. Therefore the local NDP loss $\mathcal{L}_i$ (described in Equation~\ref{eq:local-loss}) for the IL case: 
\begin{equation}
    \mathcal{L}_i(\theta_l^{(i)}) = ||F(\phi(s_t; \theta_l^{(i)}))  -  \tau_\text{demo}^{(i)}||^2
\end{equation}
For simplicity, both local and global NDPs are set to be Guassian with a fixed variance. The KL-divergence in the extra loss term to the local NDP loss (described in Equation~\ref{eq:local-loss}) therefore simply becomes: 
\begin{equation}
    D_{KL}(\pi_{l}^{(i)} || \pi_g)  = \alpha_i ||F(\phi(s_t^{(i)}; \theta_l^{(i)}))  - F(\phi(s_t^{(i)}; \theta_g))||_2 
\end{equation}
Here, let Where $s_t^{(i)}$ be the observation received by the agent while performing task $i$. Naively using a constant $\alpha$ might make the local NDPs worse. For instance, at the beginning of training, the global NDP may not be successful for every task region. Instead, we deploy the trained global policy to collect $F(\phi(s_t^{(i)}; \theta_g))$, a trajectory for every local region $i$. We set $\alpha_i = 1$ only if $F(\phi(s_t^{(i)}; \theta_g))$ is judged successful by a human. Otherwise, we set it to 0. Finally, $\mathcal{L}_g(\theta_g)$, the loss function for the global NDP is simply the imitation learning loss on the original demonstrations: 
\begin{equation}
    \mathcal{L}_g(\theta_g) = \sum_{i} ||F(\phi(s_t^{(i)}; \theta_g)) -  \tau_\text{demo}^{(i)}||^2
\end{equation}

\subsection{\ours for Reinforcement Learning}
In the RL framework, the objective is to learn a policy $\pi(a_t | s_t)$ that maximizes the sum of expected rewards \mbox{$R_t = \mathbb{E}[ \sum_{i = t}^T \gamma^{i} R(s_i, a_i, s_{i + 1})]$}. It can be difficult for RL policies to work in highly dynamic environments, especially when there is a high amount of stochasticity or variation in the task.  

In the RL setting, similarly to the imitation learning setting, we split the task into $i$ regions. Each local NDP is an RL policy and we use $\mathcal{L}_i = J_i$, where $J_i$ is the surrogate policy gradient loss from an off-the-shelf policy optimization algorithm. Specifically, we use the loss from Proximal Policy Optimization (PPO \cite{ppo}), and update the parameters of $\pi_i^{(l)}$ with respect to $\nabla J_i$.  Similarly, $\pi_g$ is optimized via PPO on the loss $\mathcal{L}_g = J_g$. To avoid divergence of the global NDP, we compute the KL divergence from the global NDP to each local NDP, as in  Equation~\ref{eq:local-loss}, and add to the local NDP training loss, $\mathcal{L}_i$. Since we use Gaussian policies with learned variance, then we have that: $D_{KL}(\pi_{l}^{(i)} || \pi_g) = \log{\frac{\sigma_g}{\sigma_i}} + \frac{\sigma_i^2 + (\mu_i - \mu_g)^2}{2\sigma_g^2} - \frac{1}{2}$.
Here the output of $\pi_{i}^{(l)}$ is $\mathcal{N}(\mu_i, \sigma_i^2)$ and that of  $\pi_g$ is $\mathcal{N}(\mu_g, \sigma_g^2)$. In practice, we found that setting $\alpha_i$ to either 0 or a very low value worked much better. An explanation for this phenomenon is that NDPs already contain more structure via the embedded dynamical system and hence do need KL-divergence constraint. The DMP parameter space these policies are in is already a lot more meaningful than the general neural network space. 

\begin{algorithm}[t]
   	\footnotesize
   	\caption{Training \ours}
   	\begin{algorithmic}
    \REQUIRE NDP Policy randomly initialized global policy $\pi_g$ with weights $\theta_g$, $M$ local regions $R_i$, for each region $i$ a local NDP $\pi_l^{(i)}$ and corresponding NN weights $\theta_l^{(i)}$, initialize empty $\mathcal{D}$
    \FOR{$1, 2, ...$ iterations}
        \FOR {$i = 1...m$}
            \STATE Run policy $\pi_l^{(i)}$ on environment $R_i$ for $H$ steps 
            \STATE Collect trajectory $F(\phi(.; \theta_l^{(i)}))$ and store into $\mathcal{D}$ 
            \STATE Compute $\mathcal{L}_{\text{local}} = \mathcal{L}_i(\theta_l^{(i)}) + \alpha_i D_{KL}(\pi_{l}^{(i)} || \pi_g)$
            \STATE $\theta_l^{(i)}  \leftarrow \theta_l^{(i)} - \eta \nabla_{\theta_l^{(i)}} \mathcal{L}_{\text{local}}$ (until convergence)
        \ENDFOR
        \STATE Compute $\mathcal{L}_{\text{BC}} = \sum_{i = 1}^{i = M} ||F(\phi(s_t; \theta_g)) -  F(\phi(s_t; \theta_l^{(i)}))||_2$
        \STATE Compute loss $\mathcal{L}_{\text{global}} = \mathcal{L}_\text{BC} + \mathcal{L}_g(\theta_g)$
        \STATE $\theta_g  \leftarrow \theta_g - \eta \nabla_{\theta_g} \mathcal{L}_{\text{global}}$ (until convergence)
    \ENDFOR
   	\end{algorithmic}
   	\label{summary}
\end{algorithm}

\begin{figure}[t!]
\vspace{0.1in}
\centering
\begin{subfigure}[b]{0.49\linewidth}
    \includegraphics[width=\linewidth]{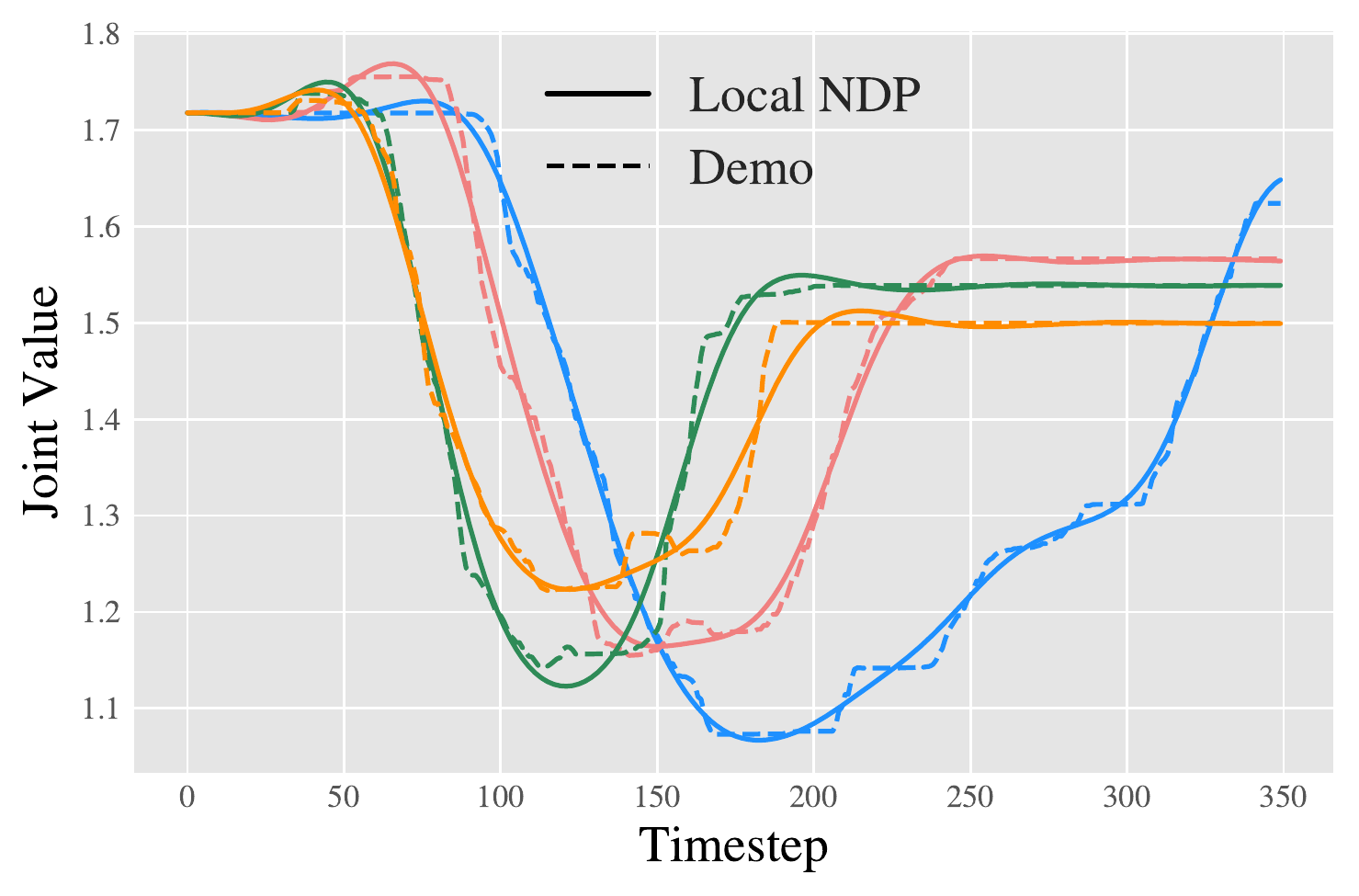}
    \vspace{-0.1in}
    \label{fig:local-ndp-0}
\end{subfigure}
\begin{subfigure}[b]{0.49\linewidth}
    \includegraphics[width=\linewidth]{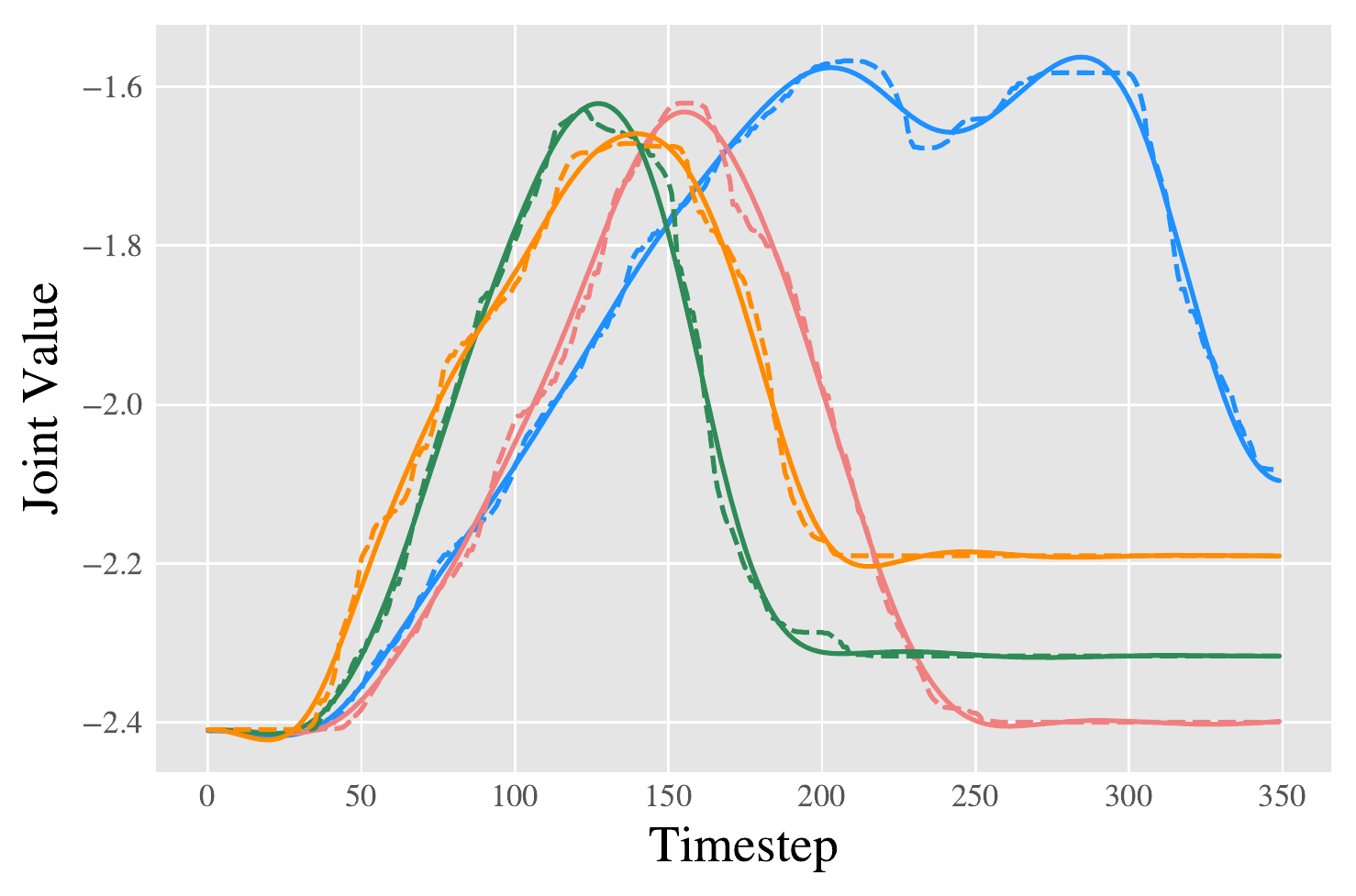}
    \vspace{-0.1in}
    \label{fig:local-ndp-2}
\end{subfigure}
 \vspace{-0.3in}
\caption{\small Visualizations of the original demonstrations and the trained local NDP on selected joints, for the real-world scooping task. The x-axis is the timestep and y is the joint value. Each curve representations a different demonstration. We can see that local NDPs can efficiently capture the desired motion in a smooth manner.}
\label{fig:local-ndp-viz}
\end{figure}

\subsection{Advantages of \ours for Real-World Robotics}
Due to their search over a physically smooth space, \ours provide safer and more efficient learning. This can be a benefit in the real world, where hardware setups can be brittle and exploration can be dangerous. In fact, in Figure~\ref{fig:local-ndp-viz} we show the trajectory for multiple joints of sampled demonstrations and the output of the corresponding fitted local NDP, which learns a much smoother version of the demonstration. 

Secondly, since \ours operate at a trajectory level, the policy $\pi$ is only used every $k$ steps. Hence fewer forward passes need to be taken by the policy network.  With large networks and computationally expensive hardware (such as robot controllers and cameras), more forward passes can actually be an impediment to the learning algorithm, as it cannot execute the task at a high enough frequency. Additionally, we can also sample trajectories at arbitrary lengths, and can therefore output a more compact trajectory if needed. 


\section{Experimental and Implementation Setup}
\begin{figure}[t]
\centering
\begin{subfigure}[b]{0.9\linewidth}
    \includegraphics[width=\linewidth]{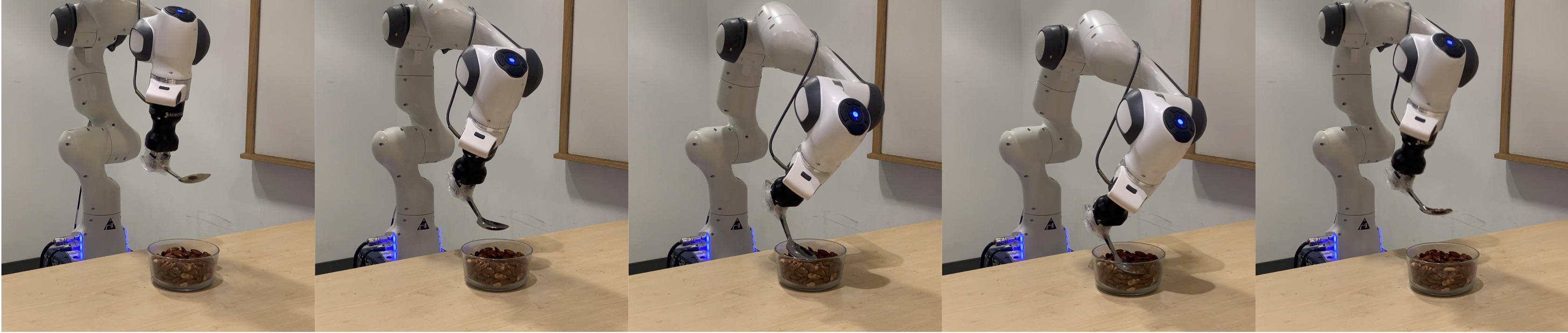}
    \vspace{-0.1in}
    \label{fig:scoop-traj}
\end{subfigure}
\begin{subfigure}[b]{0.9\linewidth}
    \includegraphics[width=\linewidth]{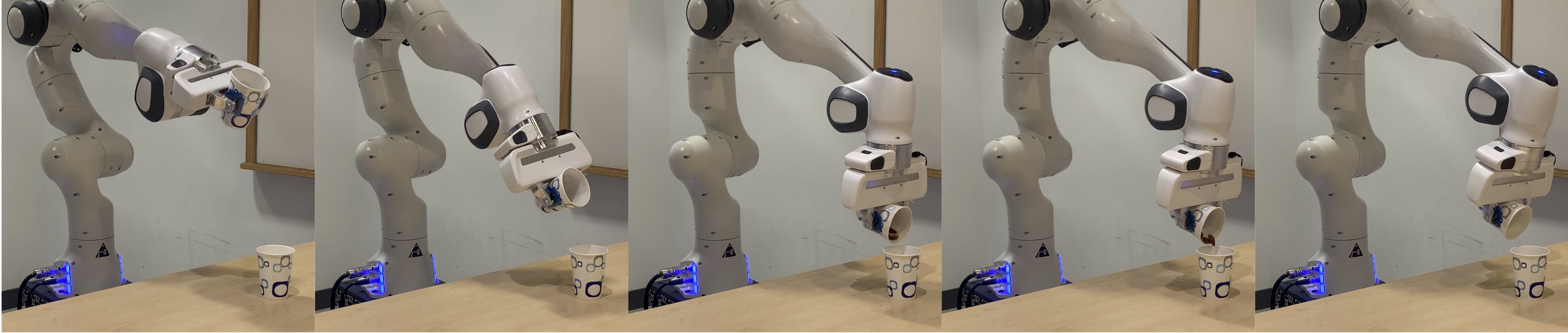}
    \vspace{-0.1in}
    \label{fig:pour-traj}
\end{subfigure}
\vspace{-0.1in}
\caption{\small Sample trajectories for Scooping (top) and Pouring (bottom) tasks, on the Franka Panda robot}
\vspace{-0.1in}
\label{fig:real-scoop-pour}
\end{figure}

\subsection{Real Robot Tasks Setup} 
For all the real world tasks (scooping, pouring and writing), we use visual inputs for the global NDP and state inputs for the local NDP. We use a Franka Panda 7 DoF robot, controlled by joint angle control. We use the robot control code from \citet{zhang2020modular}. We run the controller at about 50 Hz for both local and global policies. We are in fact bottlenecked by the frequency of the image capture and processing software. Note that a neural network policy which needs image inputs every timestep would be significantly slower (5-10Hz). An \our predicts one trajectory of $k$ steps, thus needs only one forward pass per $k$ steps allowing for a 50Hz controller. In all of our experiments, we use $k=350$. To mimic real world conditions, we vary goal locations and intentionally change the scene a little bit (slightly shift the robot, camera or the object in the robots hand). For ease of use, we utilize the same initial robot joint positions. In each of the real world tasks, a human decides whether a trial is successful or failed. More details of each task can be found in the supplementary material.

\subsection{Simulated Tasks Setup}
\begin{figure}[h]
\centering
\begin{subfigure}[b]{0.32\linewidth}
    \includegraphics[width=\linewidth]{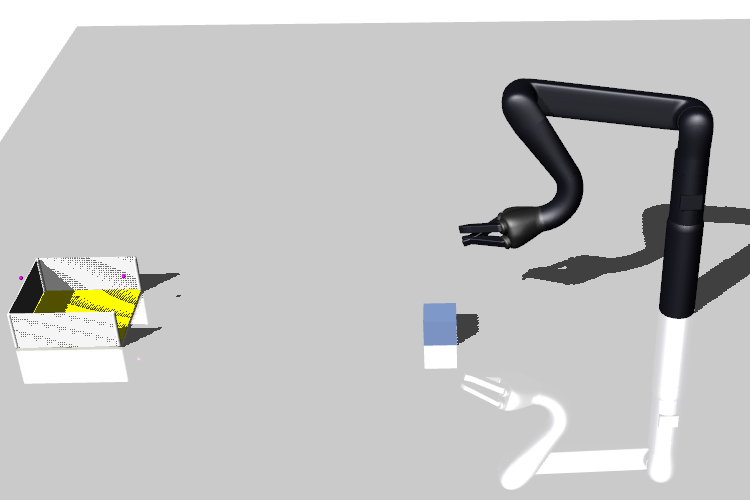}
    \vspace{-0.1in}
    \caption{\small Throwing}
    \label{fig:throw-env}
\end{subfigure}
\begin{subfigure}[b]{0.32\linewidth}
    \includegraphics[width=\linewidth]{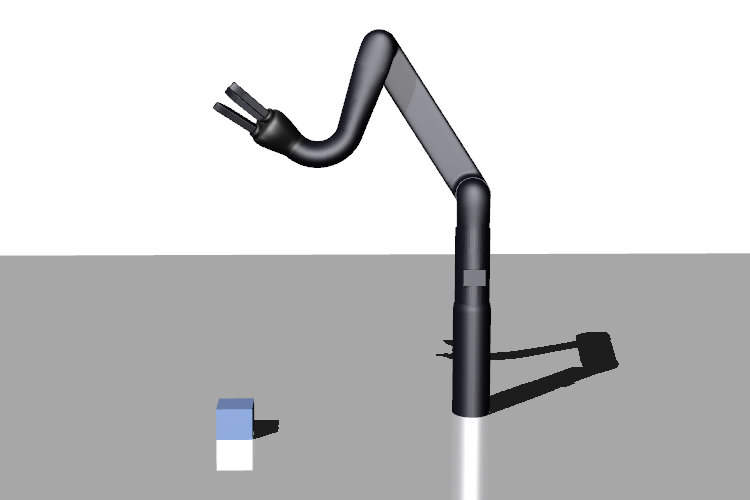}
    \vspace{-0.1in}
    \caption{\small Picking}
    \label{fig:pick-env}
\end{subfigure}
\begin{subfigure}[b]{0.32\linewidth}
    \includegraphics[width=\linewidth]{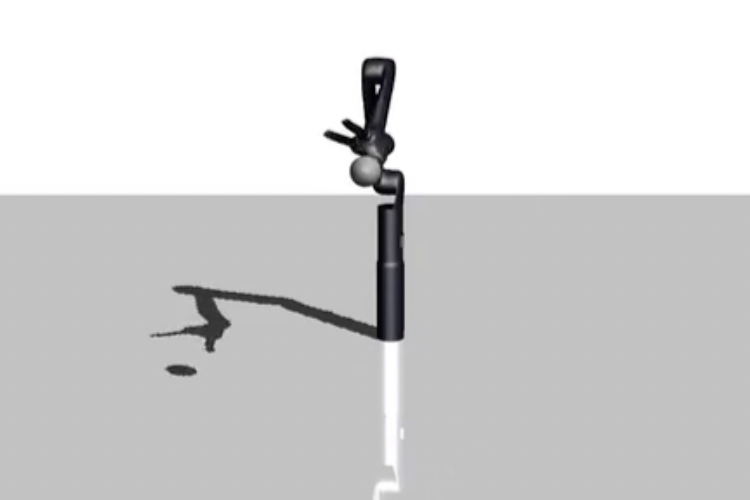}
    \vspace{-0.1in}
    \caption{\small Catching}
    \label{fig:push-env}
\end{subfigure}
\caption{\small Reinforcement Learning environments in MuJoCo \cite{todorov12mujoco}. }
\vspace{-0.1in}
\label{fig:rl-envs}
\end{figure}

We also perform  simulation experiments on dynamic tasks, inspired from tasks performed by \citet{ghosh2017divide}. The simulated robot is a 6 DoF Kinova Jaco, which we control in joint angle space. All tasks are simulated in the MuJoCo \cite{todorov12mujoco} framework. Throwing involves first grabbing then throwing a cube inside a target box. To add diversity to the task, we vary the location of the goal box. These constitute different regions of the task space. For picking, the goal is to grasp a cube and lift it as highly possible. Here the diversity comes from varying the starting position of the block. Finally, we perform the task of catching a ball being launched in the air. The goal is for the robot to catch the ball and keep it in its hand till the end of the episode. Here, we randomize the starting location of the ball. Images of these tasks can be seen in Figure~\ref{fig:rl-envs}.

\subsection{Policy Architecture and Network Pretraining}
\label{sec:method-pretrain}
In the RL setting, we use the same architecture as \citet{bahl2020neural} (2 hidden layers of 100 neurons). In the imitation learning setting, we use a very small fully connected neural network (one hidden layer with 40 neurons) for our local policies, and a similar Convolutional Neural Network (CNN) architecture to that of GPS \cite{levineFDA15}. We also use a spatial softmax layer \cite{levineFDA15}, which for each channel $c$ $f_{cx} = \sum_{ij} s_{cij} x_{ij}$ and $f_{cy} = \sum_{ij} s_{cij}y_{ij}$, where $i, j$ are pixel coordinates, and $s_{cij}$ is the spatial softmax function for pixel $a_{cij}$. We then concatenate robot joint poses to this network, and pass them through two fully connected layers and output desired joint angles. 

In order to provide the global policy with basic visual features, we pretrain the network to predict object pose data (this form of pretraining is common in robotics, e.g. in \citet{levineFDA15}). For our scooping and pouring task, we provide a similar form of pretraining, although with approximate poses. We do not use any AR markers for estimation. Instead we sample and move the robot to a position, place the object there and capture a training image. This naturally leads to imprecision in the training data, but is more realistic. For the digit writing task, we pretrain on an MNIST-like classification task, where we use a few digits written on a board by a human. 


\section{Results: H-NDPs for Imitation Learning}
We evaluate \ours on three real world tasks for imitation learning from images: Digit Writing, Scooping and Pouring. Videos at~\url{https://shikharbahl.github.io/hierarchical-ndps/} and in supplementary. One of the main focus of this work is thorough scientific evaluation in the real world itself. This experimentation involved hundreds of hours of interaction that took several weeks on hardware to complete the real-world evaluation shown in Table~\ref{tab:il-table} and Figure~\ref{fig:robot-results}. We clearly separate training and testing scenarios for each of the tasks and describe them in the following subsections as well as in the appendix.

The goal of this empirical study is to answer following scientific questions across all the tasks:
\begin{itemize}
    \item How much does the structure of dynamical systems contribute to the performance of \ours?
    \item How much does iterative refinement of global policy contribute to the performance of \ours?
    \item How much does the local-to-global structure contribute to the performance of \ours?
\end{itemize}
We attempt to answer these questions by running baseline methods. Firstly, to understand the importance of dynamical systems in \ours, we run comparisons against a method that uses iterative refinement as well as well as a local-to-global structure, but with fully connected neural network layers instead of embedded dynamical systems. This method is our implementation of GPS \cite{levineFDA15}. For every other baseline, we also design a version of that baseline with only fully connected layers, calling it vanilla NN. Secondly, to address the question of the effect of iterative refinement, we train \ours and GPS for only one iteration. To have a fair comparison, we train these baselines with 5x more demonstrations so as to provide effectively the same amount of data. However, note that these baselines have same number of interactions \textit{but have 5x more supervision} because \ours does not need more expert demonstrations after the first iteration. Finally, we test the importance of the local-to-global structure by introducing baselines that do not use it. NDP is the global policy that just learns from demonstrations. which is very similar to the method from \citet{bahl2020neural}. Vanilla NN is the fully connected counterpart of NDP.

\subsection{Task 1: Digit Writing on the Whiteboard from Image Input}
\label{sec:digit}
The goal in this task is for the robot to draw a digit on a whiteboard, given an image of the digit (ranging from 0 to 9). A dry-erase marker is attached to the robot hand. We collect 10 kinesthetic demonstrations (one for each digit 0-9) for training. We keep 10 digit images held-out which are not shown to the robot during training on which we compute the "test" success. We pretrain the global policy network using the procedure discussed in Section~\ref{sec:method-pretrain}. 

\vspace{0.6em}\noindent\textbf{Role of Dynamical System Structure:} In Table~\ref{tab:il-table}, we can see that \ours achieve the highest test success rate, 80\%. Our approach outperforms the GPS baseline by a large margin. We show a sample of the test results in Figure~\ref{fig:digit}. The picture on the left is what the robot sees at test time, and the on the right is the final output. Compared to all the other methods, \ours have the smoothest and most accurate result. Figure~\ref{fig:digit-gndp} shows a much smoother output by our method compared to that of GPS (Figure~\ref{fig:digit-gps}). The major difference between the implementation of the two approaches is that GPS uses fully connected layers instead of dynamical-system based layers. Interestingly, when comparing all other baselines (no local-to-global structure, no iterative refinement, etc) the dynamical system based methods (in Table~\ref{tab:il-table} these are \ours and NDP) all outperform their fully connected counterparts. This clearly indicates that the role of dynamical system-based structure is crucial for writing.  

\begin{figure*}[t!]
\centering
\begin{subfigure}[b]{0.32\linewidth}
    \includegraphics[width=\linewidth]{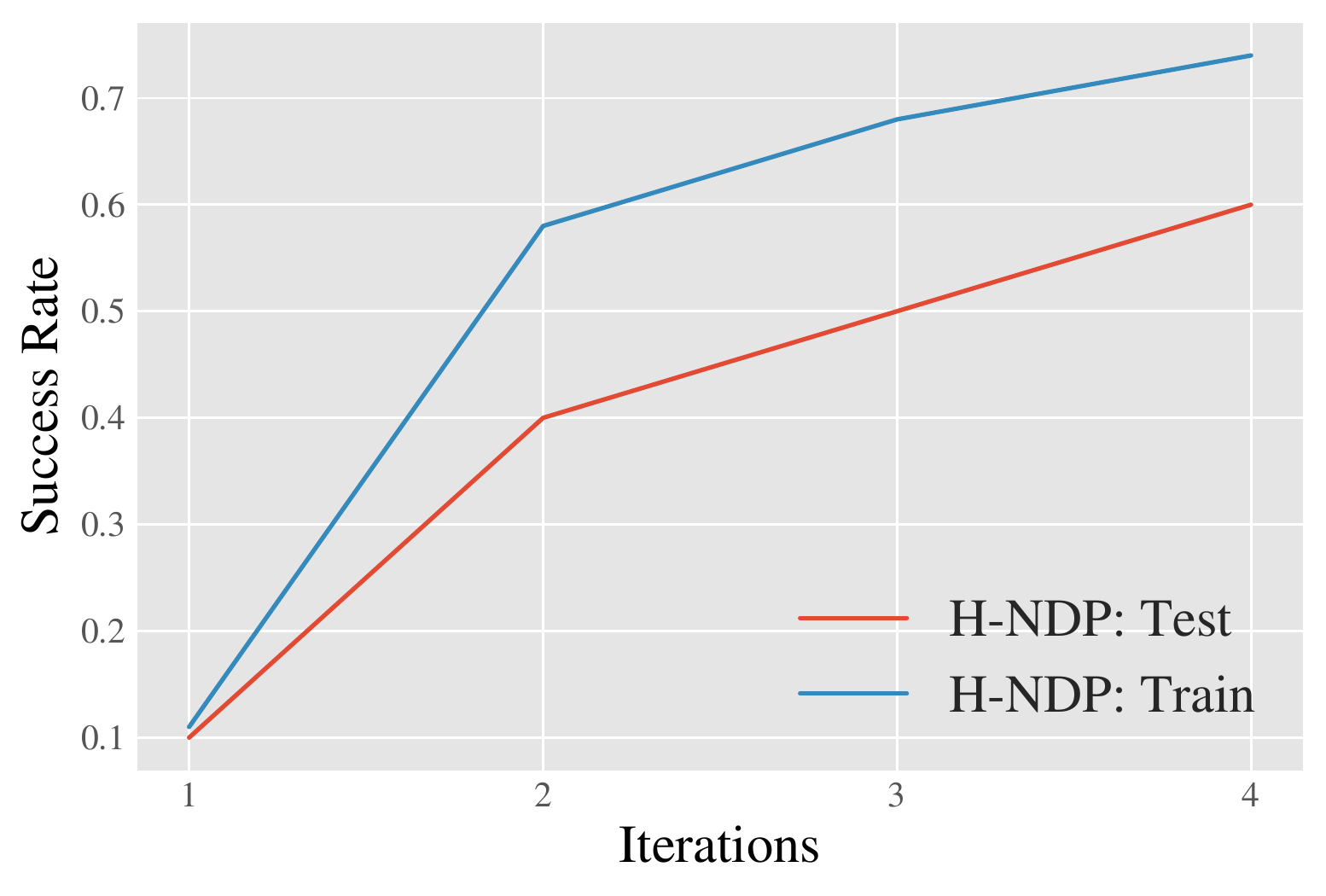}
    \vspace{-0.25in}
    \caption{\small Scooping}
    \label{fig:results-scoop}
\end{subfigure}
\begin{subfigure}[b]{0.32\linewidth}
    \includegraphics[width=\linewidth]{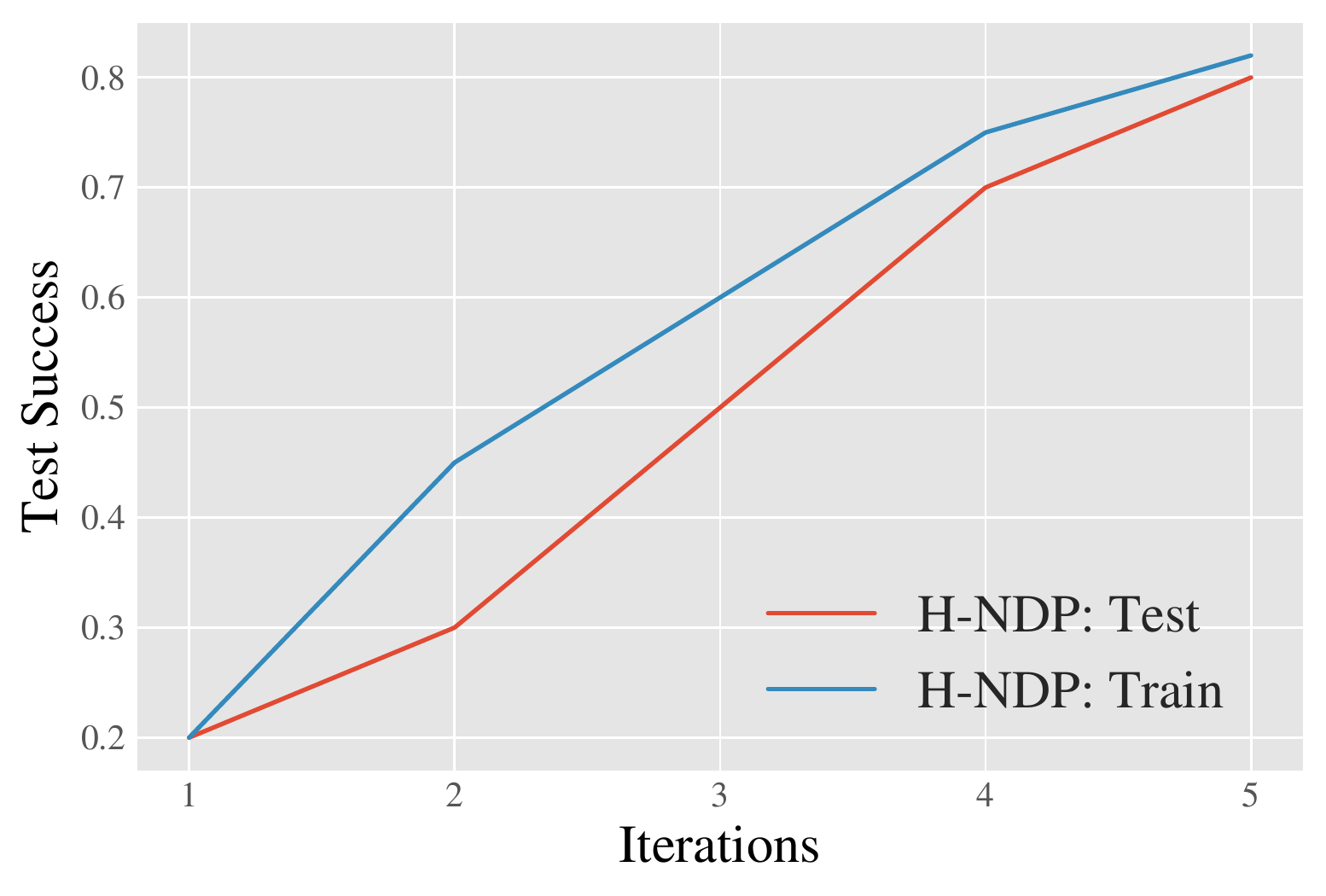}
    \vspace{-0.25in}
    \caption{\small Digit Writing}
    \label{fig:results-write}
\end{subfigure}
\begin{subfigure}[b]{0.32\linewidth}
    \includegraphics[width=\linewidth]{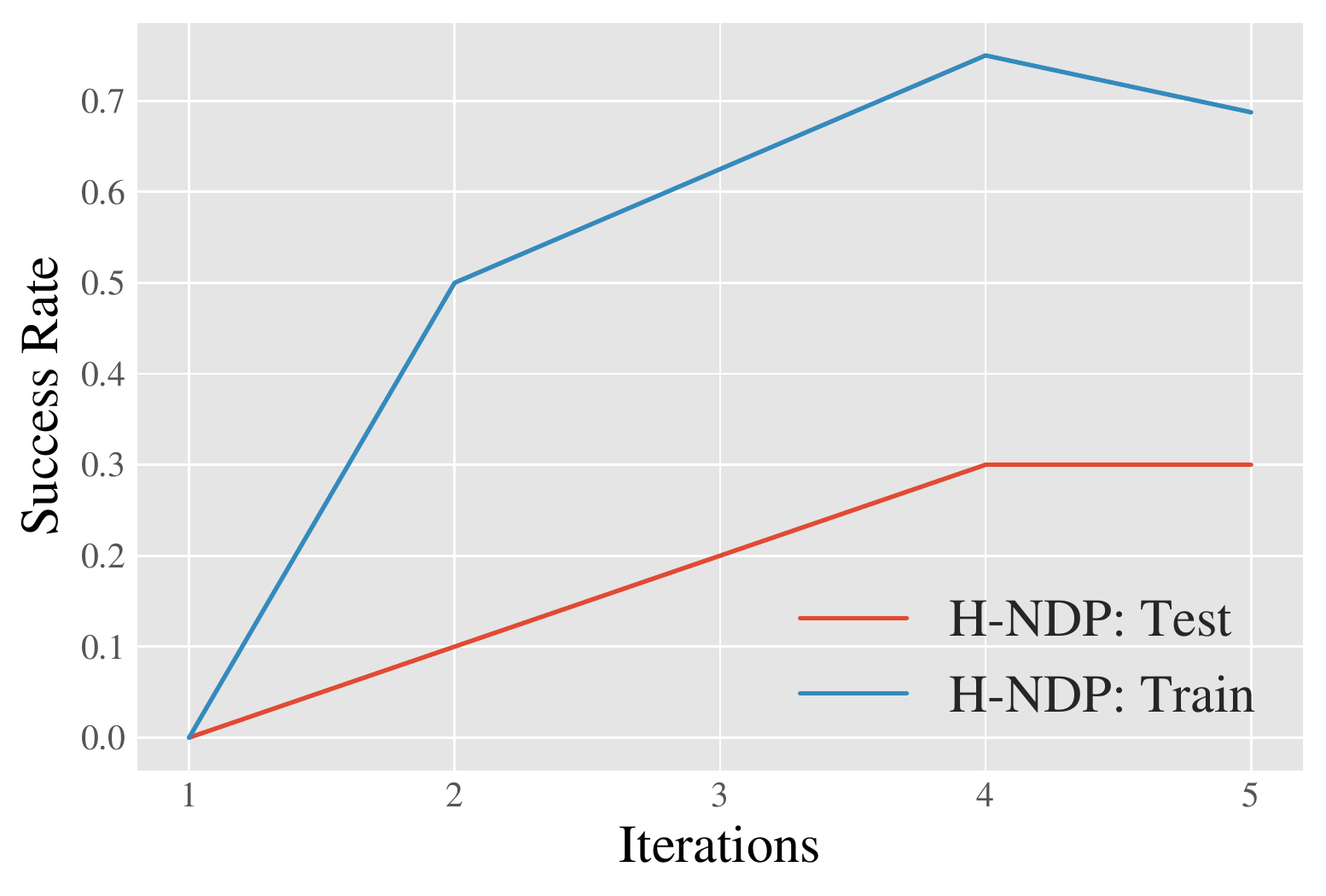}
    \vspace{-0.25in}
    \caption{\small Pouring}
    \label{fig:results-pour}
\end{subfigure}
\vspace{-0.04in}
\caption{\small Success rate for the three real-world tasks across iterations. Note that more iterations of the \our method in fact does help in learning, both in the train and test (held-out/unseen) scenarios. }
\vspace{-0.06in}
\label{fig:robot-results}
\end{figure*}

\begin{figure*}[t]
\centering
\begin{subfigure}[b]{0.42\linewidth}
    \includegraphics[width=\linewidth]{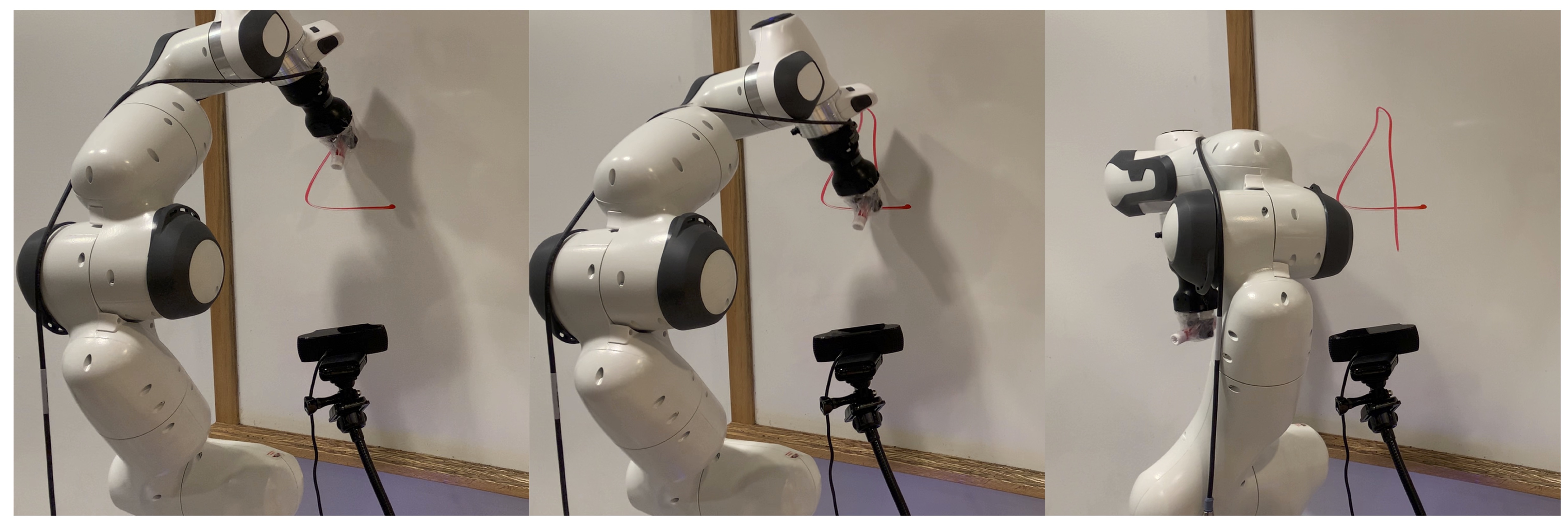}
    \vspace{-0.13in}
    \caption{\small Writing Task Setup}
    \label{fig:write-setup}
\end{subfigure}
\begin{subfigure}[b]{0.115\linewidth}
    \includegraphics[width=\linewidth]{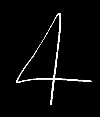}
    \vspace{-0.13in}
    \caption{\small Input}
    \label{fig:digit-img}
\end{subfigure}
\begin{subfigure}[b]{0.115\linewidth}
    \includegraphics[width=\linewidth]{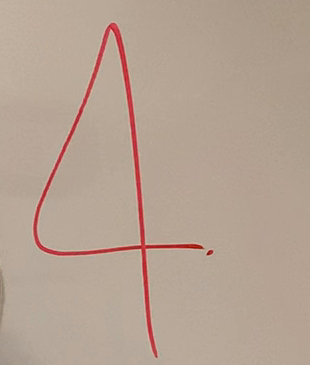}
    \vspace{-0.13in}
    \caption{\small Ours}
    \label{fig:digit-gndp}
\end{subfigure}
\begin{subfigure}[b]{0.115\linewidth}
    \includegraphics[width=\linewidth]{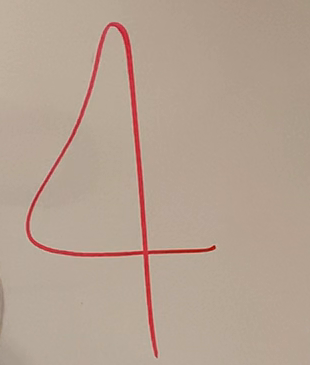}
    \vspace{-0.13in}
    \caption{\small Local NDP}
    \label{fig:digit-lndp}
\end{subfigure}
\begin{subfigure}[b]{0.115\linewidth}
    \includegraphics[width=\linewidth]{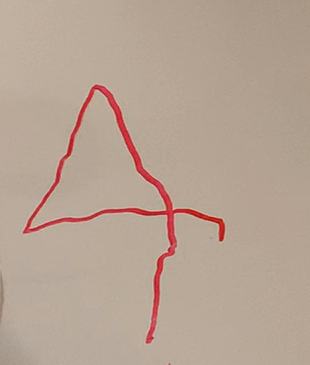}
    \vspace{-0.13in}
    \caption{\small GPS}
    \label{fig:digit-gps}
\end{subfigure}
\vspace{-0.04in}
\caption{\small Images showing the writing task setup. (a) shows the robot setup we used, a 7 DoF robot with a marker placed inside its end-effector, controlled via joint angles. (b) shows the input image (unseen at training time). (c) shows the output of our method, (d) shows the output of the local controller and (e) shows the final output of GPS \cite{levineFDA15}. We can see that our method produces a smooth and correct-looking 4.}
\vspace{-0.05in}
\label{fig:digit}
\end{figure*}

\vspace{0.6em}\noindent\textbf{Role of Iterative Refinement:} We can see that in Table~\ref{tab:il-table} that performance of \ours drops without iterative refinement (with 5x more supervision performance still drops to about 50\%). From Figure~\ref{fig:results-write}, it is clear that our method benefits from iterative refinement, as the test and train success rates increase. Interestingly, at test time our method was able to capture the "4" digit better than at training time. Despite a drop in performance, \ours without iterative refinement still outperforms almost all other baselines. 

\vspace{0.6em}\noindent\textbf{Role of Local-to-Global:} \ours clearly outperform methods that do not employ a local-to-global structure. Both the NDP and Vanilla NN baselines perform significantly worse. This is true for methods that use 1x the demos as \ours as well 5x. In fact, methods that use 1x the demos tended to fit to one demonstration and ignore the rest; i.e. Vanilla NN only output 8's for all ten digits.

\begin{table}[t]
\centering
\resizebox{\linewidth}{!}{%
\begin{tabular}{lccccc}
\toprule
 & \#Demos & \#Iter & Writing & Scooping & Pouring \\
\midrule
\multicolumn{6}{l}{\textit{No local-to-global structure}:}\vspace{0.4em}\\
NDP~\cite{bahl2020neural} & 1x & 1 & 0.2 & 0.2 & 0.0 \\
Vanilla NN & 1x & 1 & 0.1 & 0.0 & 0.0 \\
\midrule
\multicolumn{6}{l}{\textit{No local-to-global structure with 5x Demos}:}\vspace{0.4em}\\
NDP~\cite{bahl2020neural} & 5x & 1 & 0.5 & 0.3 & 0.0  \\
Vanilla NN & 5x & 1 & 0.1 & 0.0  & 0.0 \\
\midrule
\multicolumn{6}{l}{\textit{Local-to-global but no iterative refinement}:}\vspace{0.4em}\\
GPS~\cite{levineFDA15} & 5x & 1 & 0.1 & 0.0 & 0.0 \\
\ours (ours) & 5x & 1 & 0.4 & 0.3 & 0.0 \\
\midrule
\multicolumn{6}{l}{\textit{Both local-to-global and iterative refinement}:}\vspace{0.4em}\\
GPS~\cite{levineFDA15} & 1x & 5 & 0.3 & 0.0 & 0.2 \\
\midrule
\textbf{\ours (ours)} & 1x & 5 & \textbf{0.8} & \textbf{0.6} & \textbf{0.3}\\
\bottomrule
\end{tabular}}
\caption{\small Final results on the three real world tasks. We average the test success rate normalized to [$0-1$] over 10 trials on held-out testing images/locations. We compare against vanilla NDP \cite{bahl2020neural}, vanilla NN imitation, and we replace NDPs in our method with vanilla neural networks (a similar method to GPS \cite{levineFDA15}). We can see that our method outperforms all the baselines substantially.}
\label{tab:il-table}
\vspace{-0.14in}
\end{table}

\subsection{Task 2: Scooping from Image Input}
\label{sec:scooping}
In the scooping task, the robot has a spoon attached to its effector and its goal is to scoop almonds from a bowl. We vary the bowl locations, and the robot must infer from only from raw images where it should scoop. We have 18 distinct locations on the table for training and 10 distinct locations kept held-out for testing, shown in Figure 1 in the supplementary. We collect kinesthetic demonstrations on training locations. We provide the pretraining discussed in Section~\ref{sec:method-pretrain}. Figure~\ref{fig:real-scoop-pour} shows a picture of this setup.

\vspace{0.6em}\noindent\textbf{Role of Dynamical System Structure:} Similarly to the writing task, it is clear from Table~\ref{tab:il-table} that \ours drastically outperforms all baselines without dynamical system-based structure (GPS, vanilla NN, etc). In fact, all such baselines ended up executing the mean trajectory. For example, wherever the bowl would be placed, the networks would output the same trajectory towards the center of table. This clearly shows that trajectory level prediction is hard for traditional networks, and that dynamical systems are important for this task. 

\vspace{0.6em}\noindent\textbf{Role of Iterative Refinement:} In Figure~\ref{fig:results-scoop}, we see that both training and test success rates for \ours go up as we perform more iterations of refinement. For \ours, iterative refinement doubles the performance. However, even without iterative refinement, \ours still obtains a 30\% success rate, the highest of the baselines. 

\vspace{0.6em}\noindent\textbf{Role of Local-to-Global:}  We can see that \ours obtains a higher success rate at test time than any of the baselines that do not use the local-to-global structure (1x and 5x demos both). The performance gain from 5x to 1x demonstrations is not very high. This indicates just adding more data is not as important as the global-to-local framework. While \our's success rate is 60\%, even in case of failure it would go close to the bowl but not actually scoop any almonds out.

\subsection{Task 3: Pouring from Image Input}
\label{sec:pouring}
We perform experiments on a real world visual pouring task. The robot starts with a cup of almonds in its gripper, and must pour the almonds into a target cup, without any falling out. Just like scooping, the global policy must act from camera input only, and the target cup moves around to different locations. We collect kinesthetic demonstrations for 16 training locations and keep 10 held-out locations for testing as in the case of scooping, shown in the supplementary (Figure 1). Pretraining is discussed in Section~\ref{sec:method-pretrain} and the task setup in Figure~\ref{fig:real-scoop-pour}.

\vspace{0.6em}\noindent\textbf{Role of Dynamical System Structure:} This task is inherently more difficult than the others, possibly due to the size of the cups and a lot more accuracy is needed for a successful pour. \ours, however, still outperforms every other baseline, including GPS (30\% vs 20\%). During testing we observed that in the failed trials robot would go close to the cup but miss it marginally. On the other hand, GPS completely missed the target most of the time. The other vanilla NN baselines which do not perform iterative refinement would actually produce infeasible and dangerous behavior. This shows that dynamical system is important for the pouring task. 

\vspace{0.6em}\noindent\textbf{Role of Iterative Refinement:} In Table~\ref{tab:il-table}, we can see that all of the baselines without iterative refinement have 0 success. On the other hand, Figure~\ref{fig:results-pour} shows that \ours also starts with a 0\% success rate, but improves via iterative refinement. Additionally, most of the baselines produced the same trajectory for every input. This shows that iterative refinement is very helpful, especially in more challenging tasks.  

\vspace{0.6em}\noindent\textbf{Role of Local-to-Global:} All the methods that do not use the local-to-global framework have a success rate of 0\%. Both methods that do use the local-to-global structure (as well as iterative refinement), \ours and GPS, are the only ones that achieve any success. Therefore, local-to-global structure is in fact important for the pouring task.

\begin{figure*}[t]
\centering
\begin{subfigure}[b]{0.32\linewidth}
    \includegraphics[width=\linewidth]{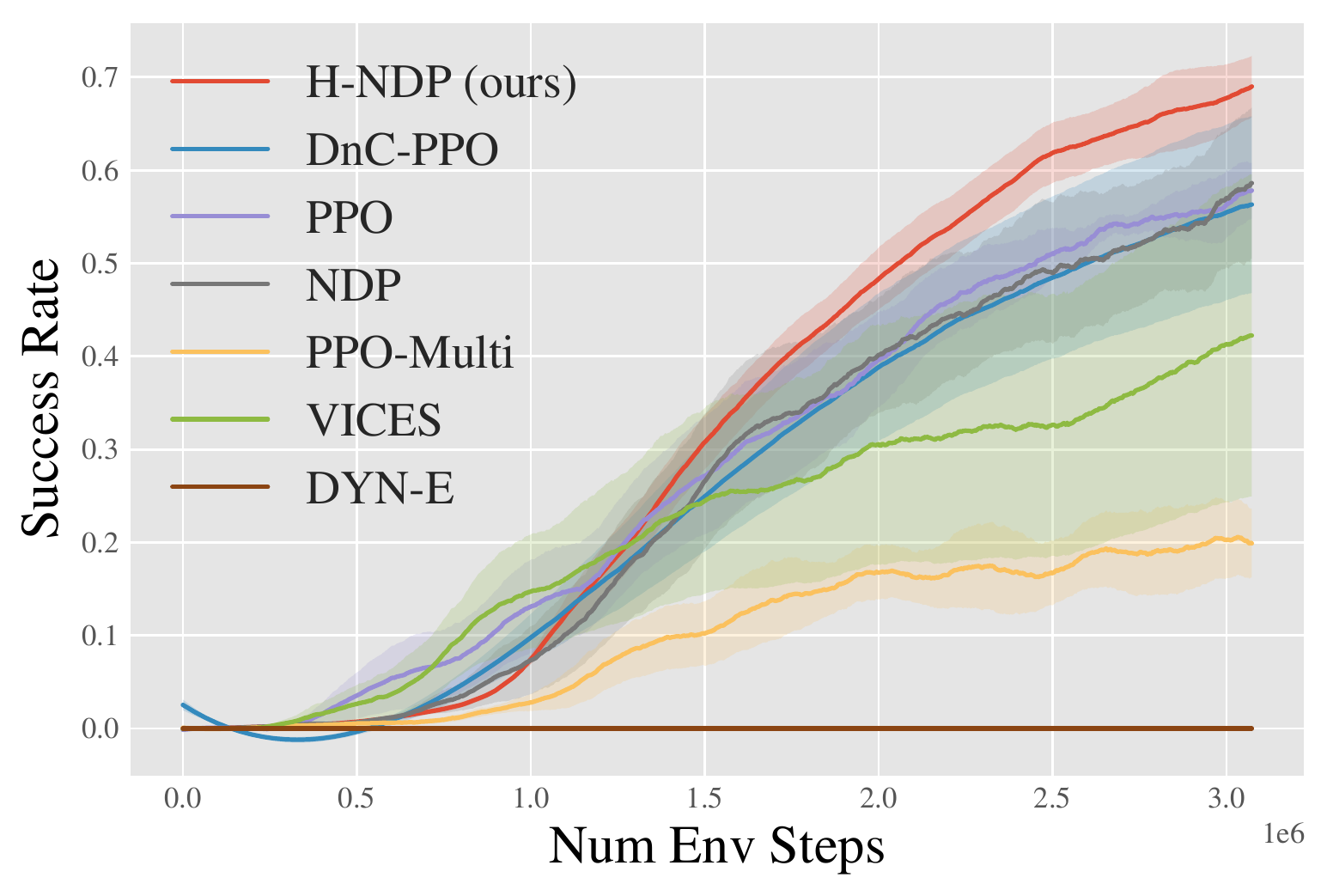}
    \vspace{-0.23in}
    \caption{\small Throwing}
    \label{fig:rl-throw}
\end{subfigure}
\begin{subfigure}[b]{0.32\linewidth}
    \includegraphics[width=\linewidth]{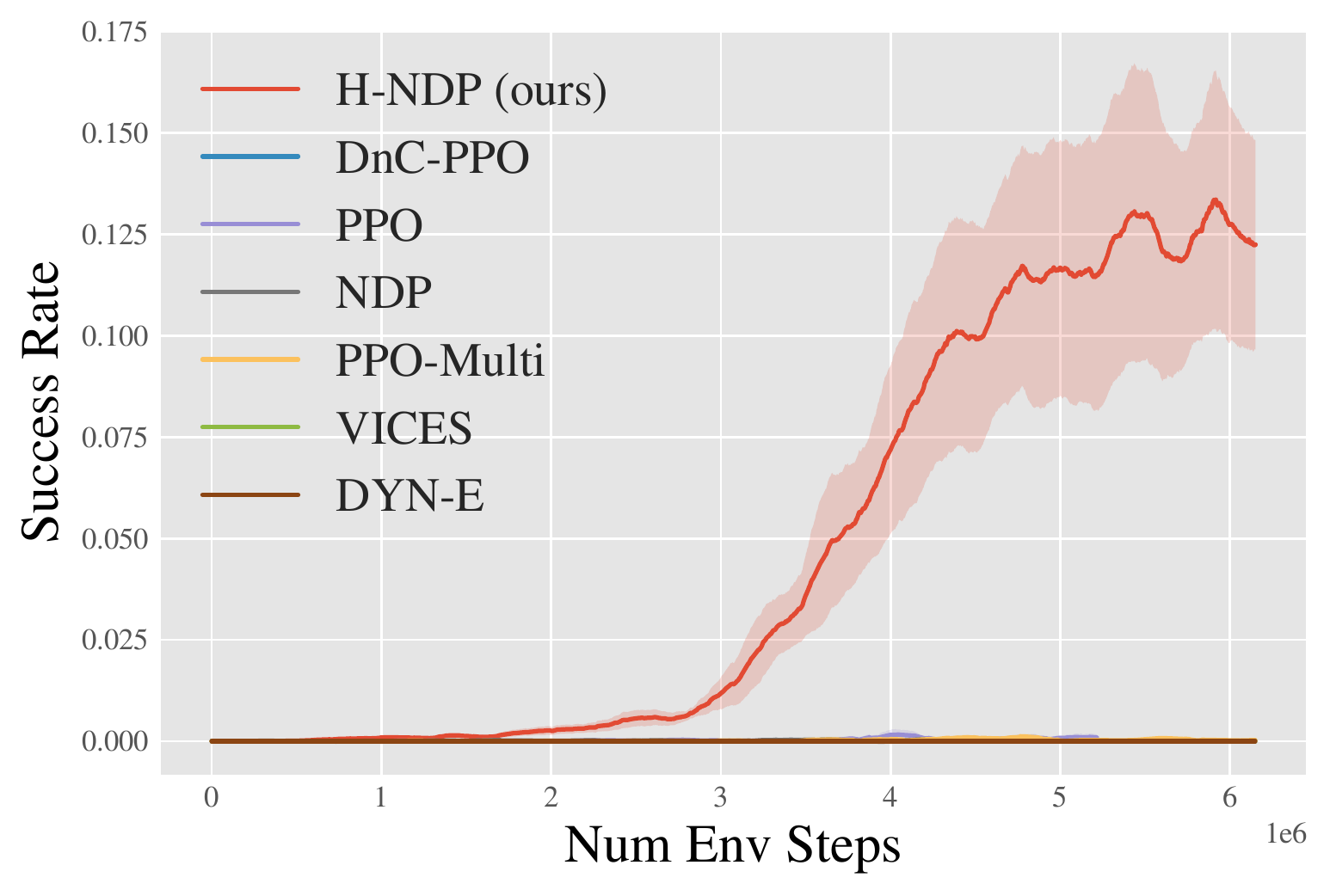}
    \vspace{-0.23in}
    \caption{\small Catching}
    \label{fig:rl-catch}
\end{subfigure}
\begin{subfigure}[b]{0.32\linewidth}
    \includegraphics[width=\linewidth]{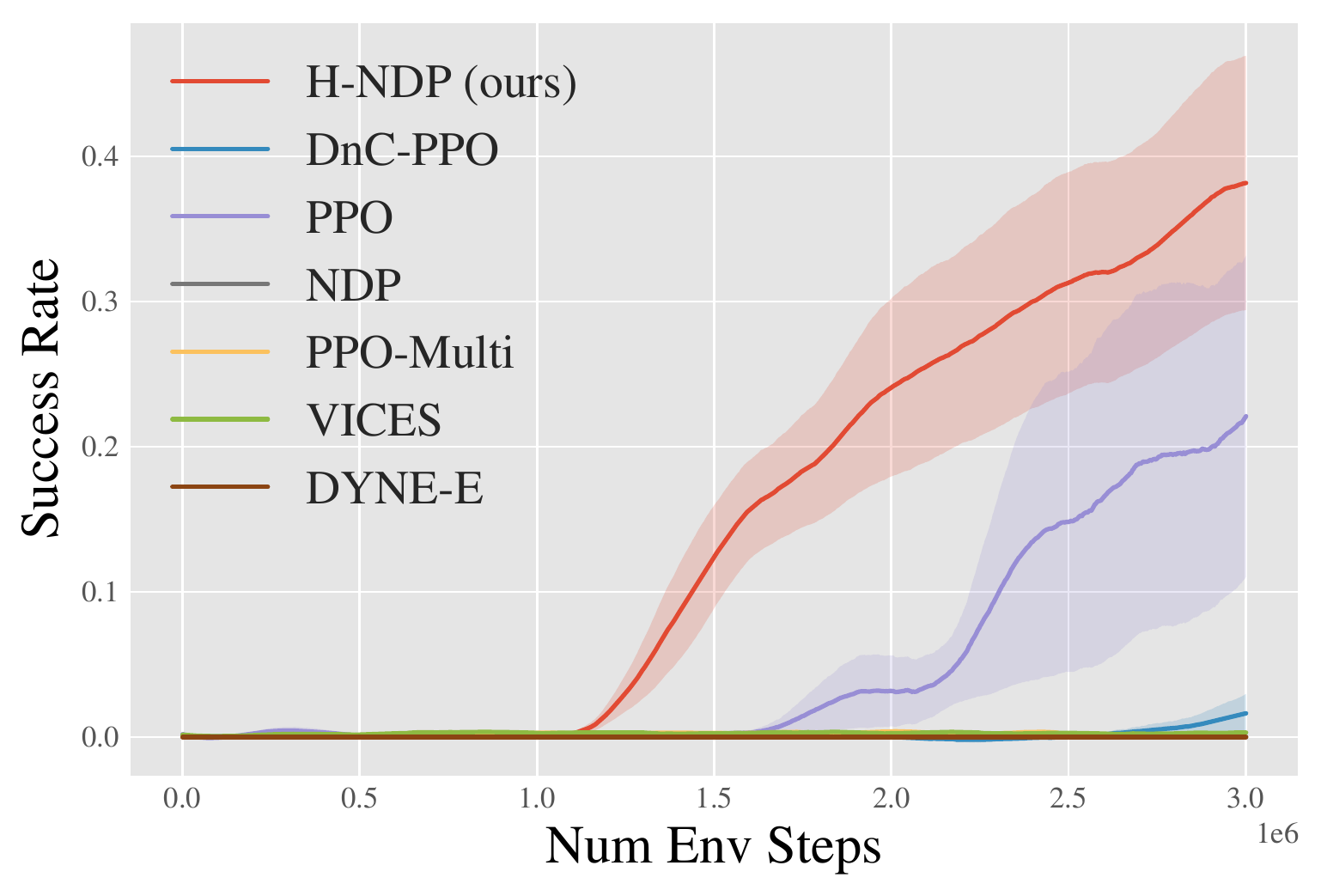}
    \vspace{-0.23in}
    \caption{\small Picking}
    \label{fig:rl-pick}
\end{subfigure}
\vspace{-0.02in}
\caption{\small Success rate for the three simulated RL tasks: throwing, catching and picking. Note that all these tasks are stochastic. Our method (red) outperforms all the baselines.}
\vspace{-0.08in}
\label{fig:rl-results}
\end{figure*}

\section{Results: \ours for Reinforcement Learning}
In the RL setting, we compare our method against several competing methods. We firstly test a similar method to \citet{ghosh2017divide} which we call PPO-DnC. This is very similar to the original method, however it uses PPO as a base algorithm, so that it can be compared apples-to-apples with \our. In another baseline, we run the NDP algorithm \cite{bahl2020neural}, equivalent to training the global NDP only. Additionally, we run our base RL algorithm, vanilla PPO \cite{ppo} as a comparison as well. To have a fair comparison, we also consider the baselines used by \citet{bahl2020neural}, in fact using their provided code. We compare against the multi-action PPO from \citet{bahl2020neural}, Variable Impedance Control in End-Effector Space (VICES) \cite{vices2019martin} and Dynamics-Aware Embeddings (DYN-E) \cite{whitney2019dynamics}. The latter two methods provide alternative parameterizations for action space: in VICES \cite{vices2019martin} the policy directly outputs parameters for a PID controller, and in DYN-E \cite{whitney2019dynamics} an action-based encoder is learnt from environment interaction.

We can see in the RL results in Figure~\ref{fig:rl-results}. We present the result of 3 random seeds run on the same codebase. We plot the success rate versus the number of environment steps taken. \our, our method, outperforms all the baselines discussed above either in terms of sample efficiency or absolute performance. This difference is especially stark for the catching task (Figure~\ref{fig:rl-catch}), since the randomness is the starting position of the ball, and even a small perturbation can have a large effect on the trajectory. \ours are able to capture this high level variation quite well, while the baselines cannot. In the other two tasks, shown in Figure~\ref{fig:rl-pick} and Figure~\ref{fig:rl-throw}, \ours are still more sample efficient and have a better final performance, even though the baselines get a relatively higher performance compared to that of catching. This is likely due to the fact that randomness in throwing and picking isn't as drastic as catching. However, overall, we can clearly see that \ours provide a strong performance boost, likely due to the embedded dynamical system which allows for smooth trajectories and efficient distillation of knowledge. 


\section{Related Work}
\label{sec:related}

\vspace{0.6em}\noindent\textbf{Robot Learning for Dynamic Tasks}
Robot learning methods have been successful for many real-world tasks. However, these tasks have been performed  in a controlled, and quasi-static setting where the robot can take arbitrarily long gap between subsequent actions~\cite{levine2016learning, agrawal2016learning,kalashnikov2018qt,pinto2015supersizing}. The real world, however, is a lot more dynamic. When humans perform daily actions, like cutting vegetables, they think at a trajectory level and not at a discrete action level. To this end, seminal work in robotics has proposed using dynamical systems to model actions and trajectories, in a continuous space. Dynamic Movement Primitives (DMPs) \cite{isprt2012dmp,schaal2006dynamic,prada2013dmp} have been widely used to perform diverse, dynamic tasks such as table tennis \cite{muelling2013tabletennis}, panckake flipping \cite{kormushev2010robot} or tether-ball \cite{parisi2015tetherball}. They are able to model smooth, natural motions, and have in fact been used to inspire many policy learning schemes \cite{conkey2019promp, Calinon2010LearningbasedCS, Calinon2016ATO, ude2010task,huang2019kmp, cheng2020rmpflow}. More recent work \cite{bahl2020neural, pahic2018deepenc, ratliff2018riemannian, chen2016dynamic} has shown DMPs can be incorporated in a differentiable, end-to-end deep learning setting, which is an attribute that \ours leverage.

\vspace{0.6em}\noindent\textbf{Hierarchical Frameworks for robot learning}
While DMPs have been used in previous works for building hierarchical policies~\cite{daniel2016hreps, stulp2012sequences, kober2009learning, pastor2011skill}, these have mostly been constrained to discrete primitives \cite{daniel2016hreps, pastor2011skill} and relatively simple settings from a perception standpoint. Previous works have also attempted to share knowledge between DMPs; for example \citet{ruckert2013learned} leverages shared basis functions for controlling multidimensional systems. To our knowledge, our work is the first to use a hierarchical local-to-global structure using DMPs. 

Hierarchical frameworks are popular for deep imitation learning setups. One prominent example is Guided Policy Search (GPS~\cite{levine2013guided,levineFDA15}) which uses a bottom-up approach by learning "expert" local controllers from state observations and then distill them into a image-based policy.
While \ours uses a similar bottom-up framework, we employ the structure of dynamical systems within our policy architecture, allowing us to perform more dynamic tasks than GPS. Furthermore, our local controllers are learnt from a few ($10-15$) demonstrations, which are a lot easier to obtain than assuming fully observable environment and hand-engineering required for GPS. 

Hierarchical learning has also long been explored in the context of RL from both top-down~\cite{vezhnevets2017feudal} as well as bottom-up~\cite{stolle2002learning, nachum2018data} perspective. \citet{ghosh2017divide} and \citet{teh2017distral} propose a hierarchical local-to-global framework to perform more complex, diverse tasks. \citet{ghosh2017divide} takes advantage of local RL policies trained via policy gradients and a global policy which imitates the former. The local-to-global interactions between neural networks can lead to suboptimal behavior, especially for more difficult dynamic tasks. On the other hand, the local-to-global interactions taking place within \ours are in a much more structured space (the space of physically plausible trajectories), leading to a stronger performance by \ours.


\section{Conclusion} 
\label{sec:conclusion}
The main contributions of this paper are:
\begin{itemize}
    \item We propose Hierarchical Neural Dynamic Policies (\ours) that embed the structure of dynamical systems in a hierarchical framework for end-to-end policy learning. H-NDPs facilitate reasoning at the trajectory level while learning from high-dimensional image inputs.
    \item We show that H-NDPs are easily integrated with standard imitation as well as reinforcement learning and achieve state-of-the-art performance across dynamic tasks in both the real world and simulation.
    \item We perform thorough scientific evaluation of held-out generalization on real-world tasks involving several hundred hours of robot interaction.
\end{itemize}

In this paper, we consider generalization only across different configurations of same objects and leave the generalization across different objects types for future work.


\section*{Acknowledgments}
We thank Vikash Kumar for fruitful discussions and grateful to Aravind Sivakumar, Russell Mendonca, Sudeep Dasari for comments on early drafts of this paper. The work was supported by NSF IIS-2024594 and AG was supported by ONR YIP.

\bibliographystyle{plainnat}
\bibliography{references}

\begin{thebibliography}{45}
\providecommand{\natexlab}[1]{#1}
\providecommand{\url}[1]{\texttt{#1}}
\expandafter\ifx\csname urlstyle\endcsname\relax
  \providecommand{\doi}[1]{doi: #1}\else
  \providecommand{\doi}{doi: \begingroup \urlstyle{rm}\Url}\fi

\bibitem[Agrawal et~al.(2016)Agrawal, Nair, Abbeel, Malik, and
  Levine]{agrawal2016learning}
Pulkit Agrawal, Ashvin Nair, Pieter Abbeel, Jitendra Malik, and Sergey Levine.
\newblock Learning to poke by poking: Experiential learning of intuitive
  physics.
\newblock \emph{NIPS}, 2016.

\bibitem[Bahl et~al.(2020)Bahl, Mukadam, Gupta, and Pathak]{bahl2020neural}
Shikhar Bahl, Mustafa Mukadam, Abhinav Gupta, and Deepak Pathak.
\newblock Neural dynamic policies for end-to-end sensorimotor learning.
\newblock In \emph{NeurIPS}, 2020.

\bibitem[Bucilua et~al.(2006)Bucilua, Caruana, and
  Niculescu-Mizil]{bucilua2006model}
Cristian Bucilua, Rich Caruana, and Alexandru Niculescu-Mizil.
\newblock Model compression.
\newblock In \emph{SIGKDD}, 2006.

\bibitem[Calinon(2016)]{Calinon2016ATO}
Sylvain Calinon.
\newblock A tutorial on task-parameterized movement learning and retrieval.
\newblock \emph{Intelligent Service Robotics}, 2016.

\bibitem[Calinon et~al.(2010)Calinon, Sardellitti, and
  Caldwell]{Calinon2010LearningbasedCS}
Sylvain Calinon, Irene Sardellitti, and Darwin~G. Caldwell.
\newblock Learning-based control strategy for safe human-robot interaction
  exploiting task and robot redundancies.
\newblock \emph{IROS}, 2010.

\bibitem[Chen et~al.(2016)Chen, Karl, and Van Der~Smagt]{chen2016dynamic}
Nutan Chen, Maximilian Karl, and Patrick Van Der~Smagt.
\newblock Dynamic movement primitives in latent space of time-dependent
  variational autoencoders.
\newblock In \emph{International Conference on Humanoid Robots (Humanoids)},
  2016.

\bibitem[Cheng et~al.(2020)Cheng, Mukadam, Issac, Birchfield, Fox, Boots, and
  Ratliff]{cheng2020rmpflow}
Ching-An Cheng, Mustafa Mukadam, Jan Issac, Stan Birchfield, Dieter Fox, Byron
  Boots, and Nathan Ratliff.
\newblock Rmpflow: A computational graph for automatic motion policy
  generation.
\newblock \emph{Algorithmic Foundations of Robotics XIII}, 2020.

\bibitem[Conkey and Hermans(2019)]{conkey2019promp}
Adam Conkey and Tucker Hermans.
\newblock Active learning of probabilistic movement primitives.
\newblock \emph{2019 IEEE-RAS 19th International Conference on Humanoid Robots
  (Humanoids)}, 2019.

\bibitem[Daniel et~al.(2016)Daniel, Neumann, Kroemer, and
  Peters]{daniel2016hreps}
Christian Daniel, Gerhard Neumann, Oliver Kroemer, and Jan Peters.
\newblock Hierarchical relative entropy policy search.
\newblock \emph{Journal of Machine Learning Research}, 2016.

\bibitem[Ghosh et~al.(2017)Ghosh, Singh, Rajeswaran, Kumar, and
  Levine]{ghosh2017divide}
Dibya Ghosh, Avi Singh, Aravind Rajeswaran, Vikash Kumar, and Sergey Levine.
\newblock Divide-and-conquer reinforcement learning.
\newblock \emph{arXiv preprint arXiv:1711.09874}, 2017.

\bibitem[Huang et~al.(2019)Huang, Rozo, Silvério, and Caldwell]{huang2019kmp}
Yanlong Huang, Leonel Rozo, João Silvério, and Darwin~G Caldwell.
\newblock Kernelized movement primitives.
\newblock \emph{The International Journal of Robotics Research}, 2019.

\bibitem[Ijspeert et~al.(2013)Ijspeert, Nakanishi, Hoffmann, Pastor, and
  Schaal]{isprt2012dmp}
Auke~Jan Ijspeert, Jun Nakanishi, Heiko Hoffmann, Peter Pastor, and Stefan
  Schaal.
\newblock Dynamical movement primitives: Learning attractor models for motor
  behaviors.
\newblock \emph{Neural Computation}, 2013.

\bibitem[Kalashnikov et~al.(2018)Kalashnikov, Irpan, Pastor, Ibarz, Herzog,
  Jang, Quillen, Holly, Kalakrishnan, Vanhoucke, et~al.]{kalashnikov2018qt}
Dmitry Kalashnikov, Alex Irpan, Peter Pastor, Julian Ibarz, Alexander Herzog,
  Eric Jang, Deirdre Quillen, Ethan Holly, Mrinal Kalakrishnan, Vincent
  Vanhoucke, et~al.
\newblock Qt-opt: Scalable deep reinforcement learning for vision-based robotic
  manipulation.
\newblock \emph{arXiv preprint arXiv:1806.10293}, 2018.

\bibitem[Kober and Peters(2009)]{kober2009learning}
Jens Kober and Jan Peters.
\newblock Learning motor primitives for robotics.
\newblock In \emph{ICRA}, 2009.

\bibitem[Kober et~al.(2011)Kober, Oztop, and Peters]{kober2011reinforcement}
Jens Kober, Erhan Oztop, and Jan Peters.
\newblock Reinforcement learning to adjust robot movements to new situations.
\newblock \emph{RSS}, 2011.

\bibitem[Kormushev et~al.(2010)Kormushev, Calinon, and
  Caldwell]{kormushev2010robot}
Petar Kormushev, Sylvain Calinon, and Darwin~G Caldwell.
\newblock Robot motor skill coordination with em-based reinforcement learning.
\newblock In \emph{IROS}, 2010.

\bibitem[Kostrikov(2018)]{pytorchrl}
Ilya Kostrikov.
\newblock Pytorch implementations of reinforcement learning algorithms.
\newblock \url{https://github.com/ikostrikov/pytorch-a2c-ppo-acktr-gail}, 2018.

\bibitem[Levine and Koltun(2013)]{levine2013guided}
Sergey Levine and Vladlen Koltun.
\newblock Guided policy search.
\newblock In \emph{ICML}, 2013.

\bibitem[Levine et~al.(2016{\natexlab{a}})Levine, Finn, Darrell, and
  Abbeel]{levineFDA15}
Sergey Levine, Chelsea Finn, Trevor Darrell, and Pieter Abbeel.
\newblock End-to-end training of deep visuomotor policies.
\newblock \emph{JMLR}, 2016{\natexlab{a}}.

\bibitem[Levine et~al.(2016{\natexlab{b}})Levine, Pastor, Krizhevsky, and
  Quillen]{levine2016learning}
Sergey Levine, Peter Pastor, Alex Krizhevsky, and Deirdre Quillen.
\newblock Learning hand-eye coordination for robotic grasping with large-scale
  data collection.
\newblock In \emph{ISER}, 2016{\natexlab{b}}.

\bibitem[Mahler et~al.(2017)Mahler, Liang, Niyaz, Laskey, Doan, Liu, Ojea, and
  Goldberg]{mahler2017dex}
Jeffrey Mahler, Jacky Liang, Sherdil Niyaz, Michael Laskey, Richard Doan, Xinyu
  Liu, Juan~Aparicio Ojea, and Ken Goldberg.
\newblock Dex-net 2.0: Deep learning to plan robust grasps with synthetic point
  clouds and analytic grasp metrics.
\newblock \emph{arXiv preprint arXiv:1703.09312}, 2017.

\bibitem[Martin-Martin et~al.(2019)Martin-Martin, Lee, Gardner, Savarese, Bohg,
  and Garg]{vices2019martin}
Roberto Martin-Martin, Michelle~A. Lee, Rachel Gardner, Silvio Savarese,
  Jeannette Bohg, and Animesh Garg.
\newblock Variable impedance control in end-effector space: An action space for
  reinforcement learning in contact-rich tasks.
\newblock \emph{IROS}, 2019.

\bibitem[Mülling et~al.(2013)Mülling, Kober, Kroemer, and
  Peters]{muelling2013tabletennis}
Katharina Mülling, Jens Kober, Oliver Kroemer, and Jan Peters.
\newblock Learning to select and generalize striking movements in robot table
  tennis.
\newblock \emph{The International Journal of Robotics Research}, 2013.

\bibitem[Nachum et~al.(2018)Nachum, Gu, Lee, and Levine]{nachum2018data}
Ofir Nachum, Shixiang Gu, Honglak Lee, and Sergey Levine.
\newblock Data-efficient hierarchical reinforcement learning.
\newblock \emph{arXiv preprint arXiv:1805.08296}, 2018.

\bibitem[Pahic et~al.(2018)Pahic, Gams, Ude, and Morimoto]{pahic2018deepenc}
Rok Pahic, Andrej Gams, Ale{\v{s}} Ude, and Jun Morimoto.
\newblock Deep encoder-decoder networks for mapping raw images to dynamic
  movement primitives.
\newblock \emph{ICRA}, 2018.

\bibitem[Parisi et~al.(2015)Parisi, Abdulsamad, Paraschos, Daniel, and
  Peters]{parisi2015tetherball}
Simone Parisi, Hany Abdulsamad, Alexandros Paraschos, Christian Daniel, and Jan
  Peters.
\newblock Reinforcement learning vs human programming in tetherball robot
  games.
\newblock In \emph{IROS}, 2015.

\bibitem[Pastor et~al.(2009)Pastor, Hoffmann, Asfour, and
  Schaal]{pastor2009motorskills}
Peter Pastor, Heiko Hoffmann, Tamim Asfour, and Stefan Schaal.
\newblock Learning and generalization of motor skills by learning from
  demonstration.
\newblock In \emph{ICRA}, 2009.

\bibitem[Pastor et~al.(2011)Pastor, Kalakrishnan, Chitta, Theodorou, and
  Schaal]{pastor2011skill}
Peter Pastor, Mrinal Kalakrishnan, Sachin Chitta, Evangelos Theodorou, and
  Stefan Schaal.
\newblock Skill learning and task outcome prediction for manipulation.
\newblock In \emph{ICRA}, 2011.

\bibitem[Pinto and Gupta(2016)]{pinto2015supersizing}
Lerrel Pinto and Abhinav Gupta.
\newblock Supersizing self-supervision: Learning to grasp from 50k tries and
  700 robot hours.
\newblock \emph{ICRA}, 2016.

\bibitem[{Prada} et~al.(2013){Prada}, {Remazeilles}, {Koene}, and
  {Endo}]{prada2013dmp}
M.~{Prada}, A.~{Remazeilles}, A.~{Koene}, and S.~{Endo}.
\newblock Dynamic movement primitives for human-robot interaction: Comparison
  with human behavioral observation.
\newblock In \emph{International Conference on Intelligent Robots and Systems},
  2013.

\bibitem[Ratliff et~al.(2018)Ratliff, Issac, Kappler, Birchfield, and
  Fox]{ratliff2018riemannian}
Nathan~D Ratliff, Jan Issac, Daniel Kappler, Stan Birchfield, and Dieter Fox.
\newblock Riemannian motion policies.
\newblock \emph{arXiv preprint arXiv:1801.02854}, 2018.

\bibitem[R{\"u}ckert and d'Avella(2013)]{ruckert2013learned}
Elmar R{\"u}ckert and Andrea d'Avella.
\newblock Learned parametrized dynamic movement primitives with shared
  synergies for controlling robotic and musculoskeletal systems.
\newblock \emph{Frontiers in computational neuroscience}, 2013.

\bibitem[Rusu et~al.(2015)Rusu, Colmenarejo, Gulcehre, Desjardins, Kirkpatrick,
  Pascanu, Mnih, Kavukcuoglu, and Hadsell]{rusu2015policy}
Andrei~A Rusu, Sergio~Gomez Colmenarejo, Caglar Gulcehre, Guillaume Desjardins,
  James Kirkpatrick, Razvan Pascanu, Volodymyr Mnih, Koray Kavukcuoglu, and
  Raia Hadsell.
\newblock Policy distillation.
\newblock \emph{arXiv preprint arXiv:1511.06295}, 2015.

\bibitem[Schaal(2006)]{schaal2006dynamic}
Stefan Schaal.
\newblock Dynamic movement primitives-a framework for motor control in humans
  and humanoid robotics.
\newblock In \emph{Adaptive motion of animals and machines}. Springer, 2006.

\bibitem[Schulman et~al.(2017)Schulman, Wolski, Dhariwal, Radford, and
  Klimov]{ppo}
John Schulman, Filip Wolski, Prafulla Dhariwal, Alec Radford, and Oleg Klimov.
\newblock Proximal policy optimization algorithms.
\newblock \emph{arXiv:1707.06347}, 2017.

\bibitem[Stolle and Precup(2002)]{stolle2002learning}
Martin Stolle and Doina Precup.
\newblock Learning options in reinforcement learning.
\newblock In \emph{International Symposium on abstraction, reformulation, and
  approximation}, 2002.

\bibitem[{Stulp} et~al.(2012){Stulp}, {Theodorou}, and
  {Schaal}]{stulp2012sequences}
F.~{Stulp}, E.~A. {Theodorou}, and S.~{Schaal}.
\newblock Reinforcement learning with sequences of motion primitives for robust
  manipulation.
\newblock \emph{Transactions on Robotics}, 2012.

\bibitem[Sun et~al.(2001)Sun, Merrill, and Peterson]{sun2001implicit}
Ron Sun, Edward Merrill, and Todd Peterson.
\newblock From implicit skills to explicit knowledge: A bottom-up model of
  skill learning.
\newblock \emph{Cognitive science}, 2001.

\bibitem[Teh et~al.(2017)Teh, Bapst, Czarnecki, Quan, Kirkpatrick, Hadsell,
  Heess, and Pascanu]{teh2017distral}
Yee~Whye Teh, Victor Bapst, Wojciech~Marian Czarnecki, John Quan, James
  Kirkpatrick, Raia Hadsell, Nicolas Heess, and Razvan Pascanu.
\newblock Distral: Robust multitask reinforcement learning.
\newblock \emph{arXiv preprint arXiv:1707.04175}, 2017.

\bibitem[Todorov et~al.(2012)Todorov, Erez, and Tassa]{todorov12mujoco}
Emanuel Todorov, Tom Erez, and Yuval Tassa.
\newblock {MuJoCo: A physics engine for model-based control}.
\newblock In \emph{The IEEE/RSJ International Conference on Intelligent Robots
  and Systems}, 2012.

\bibitem[Ude et~al.(2010)Ude, Gams, Asfour, and Morimoto]{ude2010task}
Ale{\v{s}} Ude, Andrej Gams, Tamim Asfour, and Jun Morimoto.
\newblock Task-specific generalization of discrete and periodic dynamic
  movement primitives.
\newblock \emph{Transactions on Robotics}, 2010.

\bibitem[Vezhnevets et~al.(2017)Vezhnevets, Osindero, Schaul, Heess, Jaderberg,
  Silver, and Kavukcuoglu]{vezhnevets2017feudal}
Alexander~Sasha Vezhnevets, Simon Osindero, Tom Schaul, Nicolas Heess, Max
  Jaderberg, David Silver, and Koray Kavukcuoglu.
\newblock Feudal networks for hierarchical reinforcement learning.
\newblock In \emph{ICML}, 2017.

\bibitem[Whitney et~al.(2019)Whitney, Agarwal, Cho, and
  Gupta]{whitney2019dynamics}
William Whitney, Rajat Agarwal, Kyunghyun Cho, and Abhinav Gupta.
\newblock Dynamics-aware embeddings.
\newblock \emph{arXiv preprint arXiv:1908.09357}, 2019.

\bibitem[Zhang et~al.(2020)Zhang, Sharma, Liang, and Kroemer]{zhang2020modular}
Kevin Zhang, Mohit Sharma, Jacky Liang, and Oliver Kroemer.
\newblock A modular robotic arm control stack for research: Franka-interface
  and frankapy.
\newblock \emph{arXiv preprint arXiv:2011.02398}, 2020.

\bibitem[Zhu et~al.(2020)Zhu, Wong, Mandlekar, and
  Mart\'{i}n-Mart\'{i}n]{robosuite2020}
Yuke Zhu, Josiah Wong, Ajay Mandlekar, and Roberto Mart\'{i}n-Mart\'{i}n.
\newblock robosuite: A modular simulation framework and benchmark for robot
  learning.
\newblock In \emph{arXiv preprint arXiv:2009.12293}, 2020.

\end{thebibliography}

\clearpage
\appendix
\subsection{Videos}
Videos of our results and comparison with baselines are available at~\url{https://shikharbahl.github.io/hierarchical-ndps/}. For every task, we show the original demonstration, results of GPS~\cite{levineFDA15} baseline as well as \ours results. For each of these tasks, we can see that our method is not only much more accurate than the baseline but also smoother and thus safer to operate. The video shows that the MLP based GPS baseline policy produces very shaky trajectories.

\subsection{Implementation Details}
\subsubsection{Hyper-parameters and Design Choices}

Architectures: In the imitation learning case, we use the same image and network sizes as \citet{levineFDA15}, everything up to the NDP layer, which is from \citet{bahl2020neural}. Our global policies have 3 convolutional layers, of kernel sizes 7, 5 and 5, have 64, 32 and 32 output channels. After the convolutional layers, just like \citet{levineFDA15}, we employ a spacial softmax, where we take the expected features in the x and y axis and concatenate them. We then get a flattened vector of size 64. Similarly to \citet{levineFDA15} we concatenate the robot joint positions and 3D end-effector pose into these features. We then have 2 fully connected layers of hidden size 40. We use ReLU non-linearities. For the rest of the network, we use the exact architecture from \citet{bahl2020neural}. Our local NDPs are similar to those provided by \citet{bahl2020neural}, but use a smaller hidden layer size (40 instead of 100). All images are 224 x 224, except for the ones used in the writing which are 100 x 100.

Pretraining: For the scooping and pouring tasks, we collect data from about 150-200 goal locations. We move the object to these 250 locations, and then randomly sample 5 actions for the robot take. We use all the layers of the global policy described above, save for the last layer, which we replace with a layer of output 3. We train the network to output the pose of the object. This is a similar pre-training procedure as \citet{levineFDA15}. 

Training: In our imitation learning experiments, we use the parameter $\alpha$ to weigh the contributions of the original demonstrations to the IL loss function, in the outer loop. For writing we use $\alpha = 1$, for pour and scooping we use $\alpha = 0.5$. In the inner loop, we use $\beta = 0.1$ to weigh the contributions of the outputs of the global policy. Training such networks takes about 2 hours. 

RL: For the RL experiments, we use the same exact architectures and hyperparameters for every method as \citet{bahl2020neural}, which is built on the PPO~\cite{ppo} implementation from \citet{pytorchrl}. All networks use hidden layer sizes [100, 100] with tanh non linearities. We use the code from \citet{robosuite2020} for VICES \cite{vices2019martin} and from \citet{whitney2019dynamics} for DYN-E. 

We present the hyperparameters we used for the RL experiments in Table~\ref{tab:hyperparams}.  
 
\begin{table}[ht]
\centering
\begin{tabular}{lc}
\toprule
\textbf{Hyperparameter} & \textbf{Value}\\
\midrule
Learning Rate & 0.00025 \\
Discount Factor ($\gamma$)& 0.99 \\ 
GAE Discount Factor & 0.95 \\ 
Entropy Coefficient & 0 \\ 
Normalized Observations & True \\ 
Normalized Returns & True \\ 
Value Loss Coefficient & 0.5 \\ 
Maximum Gradient Norm & 0.5 \\ 
PPO Mini-Batches & 32 \\ 
PPO Epochs & 10 \\ 
Clip Parameter & 0.1 \\ 
Optimizer & Adam \\
Batch Size & 2048 \\
RMSprop optimizer epsilon &  $10^{-5}$ \\ 
\bottomrule
\end{tabular}
\vspace{0.1cm}
\caption{Hyperparameters used by \ours for RL}
\label{tab:hyperparams}
\end{table}

\subsubsection{Real Robot Setup}

We run all of the robot experiments in the Franka Panda 7 DoF robot. We use joint angle control, and use the controller package provided by \citet{zhang2020modular}. We run the global policy at 50 Hz. Each of our trajectories are 15-17s in length and about 350 timesteps. We constrain our method to run at 50 Hz due to our perception system. We use a webcam for all of our images and a ROS communication pootocol. 
Writing: The writing task consists of taping a dry-erase whiteboard marker to the robot-endeffector, and placing the robot near a whiteboard. We collect 10 demonstrations - one for each digit by kinesthetic teaching. After fitting the local policies, the webcam is used to take a picture of the digit. Since the pen is red and the whiteboard is white, we use OpenCV to process the image and invert the colors so that it resembles an MNIST image. It is then resized to be 100x100. The 10 test images are processed in the same way but are unseen at training time. 

\begin{figure*}[t!]
\centering
\begin{subfigure}[b]{0.47\linewidth}
    \includegraphics[width=\linewidth]{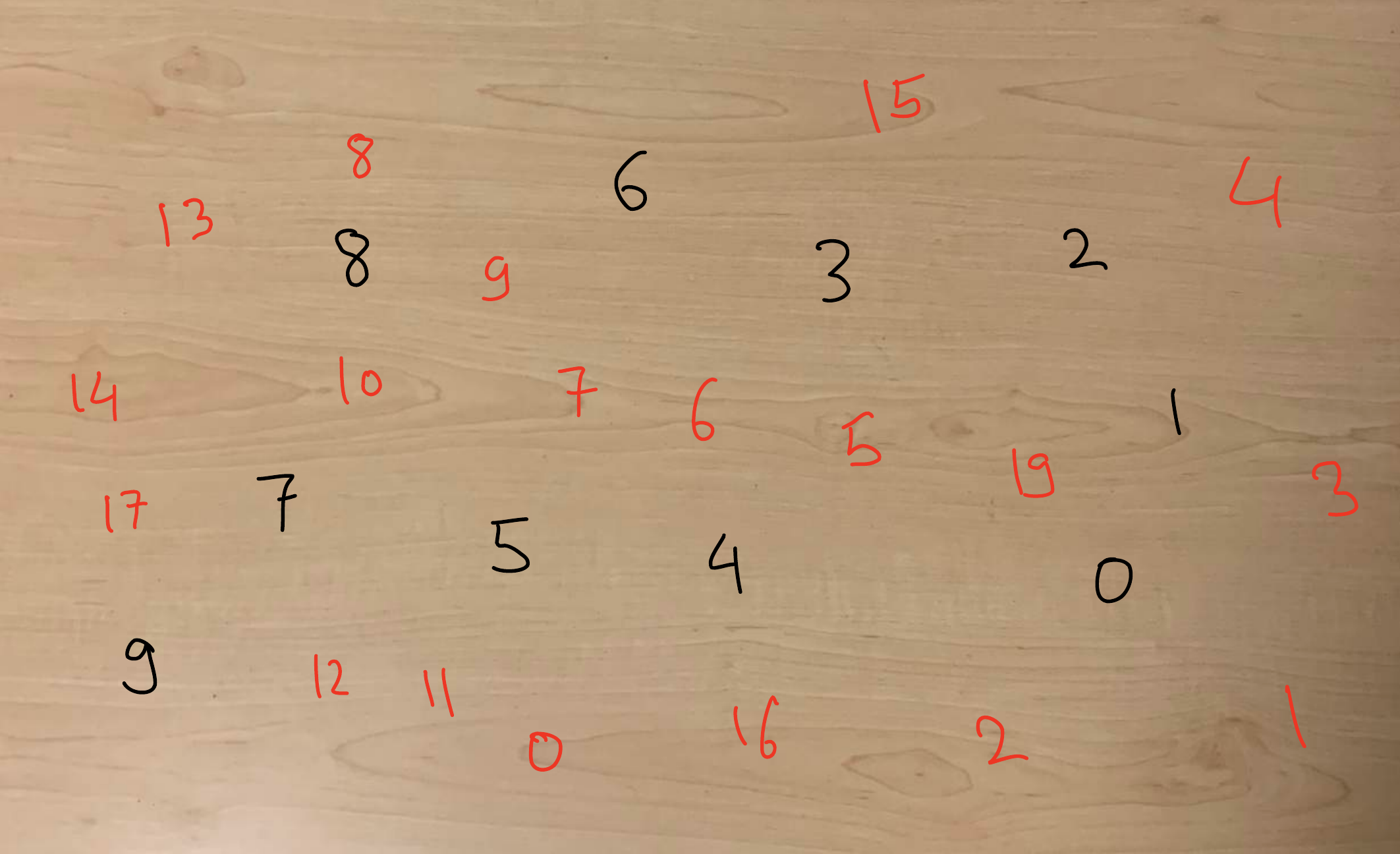}
    \caption{\small Scooping}
    \label{fig:scoop-table}
\end{subfigure}
\begin{subfigure}[b]{0.47\linewidth}
    \includegraphics[width=\linewidth]{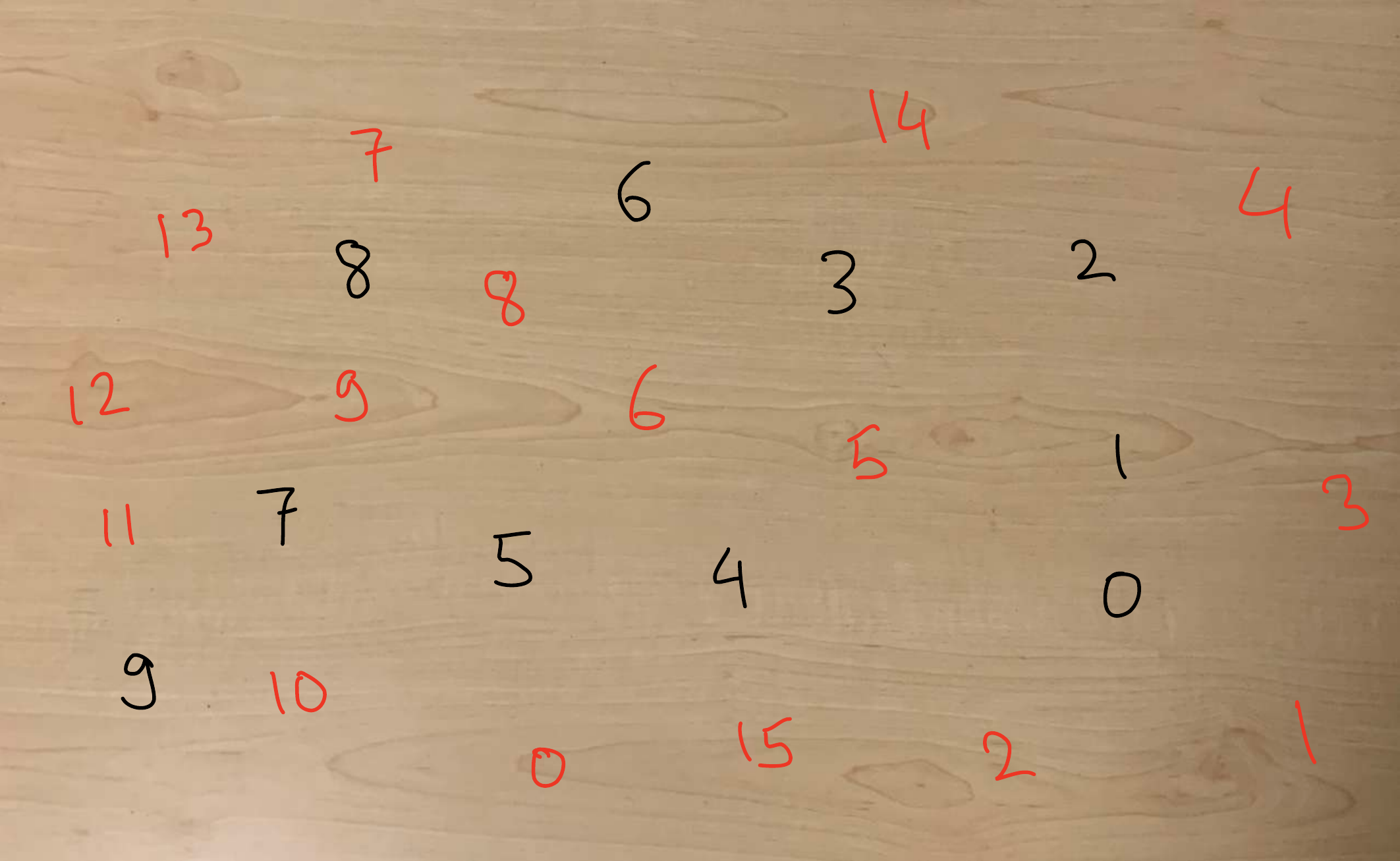}
    \caption{\small Pouring}
    \label{fig:pour-table}
\end{subfigure}
\caption{\small Here, we show the training (red) and testing (black) locations on the table for the pouring and scooping tasks. }
\vspace{-0.1in}
\end{figure*}

Scooping: The scooping task consists of a metal serving spoon attached to the end-effector. The goal is to scoop almonds from medium sized bowl. Demonstrations are collected via kinesthetic teaching, one for each table location. We collect 18 demonstrations for 18 locations each. We use 10 different test locations, which are different from the training locations. Figure~\ref{fig:scoop-table} shows the training and testing locations. The images collected by the webcam are 224 x 224.

Pouring: The pouring task consists of a 100ml paper cup attached to the end-effector. The goal is to pour the almonds in the cup into another 100ml paper cup on the table. Similarly to scooping, demonstrations are collected via kinesthetic teaching. We use 16 of the 18 training locations as the scooping task and the same test locations. Figure~\ref{fig:pour-table} shows these locations on the table. Similarly to scooping, the images we use are 224x224. 

\subsubsection{Simulated Environments}
Our simulated environments use MuJoCo \cite{todorov12mujoco}. We adapted tasks our tasks from~\citet{ghosh2017divide} (\url{https://github.com/dibyaghosh/dnc}). The original tasks use torque control, however we modified the environments since our method uses joint angle control. 

\subsubsection{Codebases}
Our RL and imitation learning experiments build upon the NDP code from \url{https://shikharbahl.github.io/neural-dynamic-policies/}, which is based on the PPO \cite{ppo} implementation \url{https://github.com/ikostrikov/pytorch-a2c-ppo-acktr-gail} \cite{pytorchrl}. Our DnC implementation is based on top of the NDP code and is inspired by the code from \citet{ghosh2017divide} (\url{https://github.com/dibyaghosh/dnc}). The VICES \cite{vices2019martin} implementation we use is from \url{https://github.com/ARISE-Initiative/robosuite} and the DYN-E \cite{whitney2019dynamics} implementation from \url{https://github.com/willwhitney/dynamics-aware-embeddings}.

\end{document}